\documentclass[journal]{IEEEtran}

\usepackage{times}
\usepackage{multicol}
\usepackage{multirow}
\usepackage{amsmath,amssymb,amsfonts}
\usepackage{algorithmic}
\usepackage{graphicx}
\usepackage{textcomp}
\usepackage{float}
\usepackage{multirow}
\usepackage{subcaption}
\usepackage{tabularx}

\usepackage{amsthm}
\usepackage{cite}
\usepackage[normalem]{ulem}
\usepackage{makecell}
\usepackage[table]{xcolor}

\definecolor{lightblue}{rgb}{0.91,0.957,0.973}

\newcommand{\stddev}[1]{\textcolor{darkgray}{\pm #1}}

\theoremstyle{definition}
\newtheorem{problem}{Problem}

\newcommand{\pref}{\mathcal{P}_{\text{ref}}}

\graphicspath{{figures/}}

\def\BibTeX{{\rm B\kern-.05em{\sc i\kern-.025em b}\kern-.08em
    T\kern-.1667em\lower.7ex\hbox{E}\kern-.125emX}}
\begin{document}

\title{A Two-Stage Optimization-based Motion Planner for Safe Urban Driving}

\author{Francisco Eiras$^{1,2}$,
Majd Hawasly$^{1}$,
Stefano V. Albrecht$^{1,3}$, 
Subramanian Ramamoorthy$^{1,3}$%
\thanks{$^{1}$FiveAI Ltd, United Kingdom, 
{\tt\small \{first.last\}@five.ai
        }}%
\thanks{$^{2}$Dept. of Engineering Science, University of Oxford, United Kingdom}
\thanks{$^{3}$School of Informatics, University of Edinburgh, United Kingdom}
}

\markboth{IEEE Transactions on Robotics,~Vol.~, No.~, ~2020}%
{Eiras \MakeLowercase{\textit{et al.}}: A Two-Stage Optimization-based Motion Planner for Safe Urban Driving}

\maketitle

\IEEEpeerreviewmaketitle

\thispagestyle{empty}
\pagestyle{empty}

\begin{abstract}
Recent road trials have shown that guaranteeing the safety of driving decisions is essential for the wider adoption of autonomous vehicle technology. One promising direction is to pose safety requirements as planning constraints in nonlinear, non-convex optimization problems of motion synthesis. However, many implementations of this approach are limited by uncertain convergence and local optimality of the solutions achieved, affecting overall robustness. To improve upon these issues, we propose a novel two-stage optimization framework: in the first stage, we find a solution to a Mixed-Integer Linear Programming (MILP) formulation of the motion synthesis problem, the output of which initializes a second Nonlinear Programming (NLP) stage. The MILP stage enforces hard constraints of safety and road rule compliance generating a solution in the right subspace, while the NLP stage refines the solution within the safety bounds for feasibility and smoothness. We demonstrate the effectiveness of our framework via simulated experiments of complex urban driving scenarios, outperforming a state-of-the-art baseline in metrics of convergence, comfort and progress.
\end{abstract}


\section{Introduction}
\begin{figure}[t]
    \centering
    \begin{subfigure}{0.48\textwidth}
        \centering
        \includegraphics[width=\textwidth]{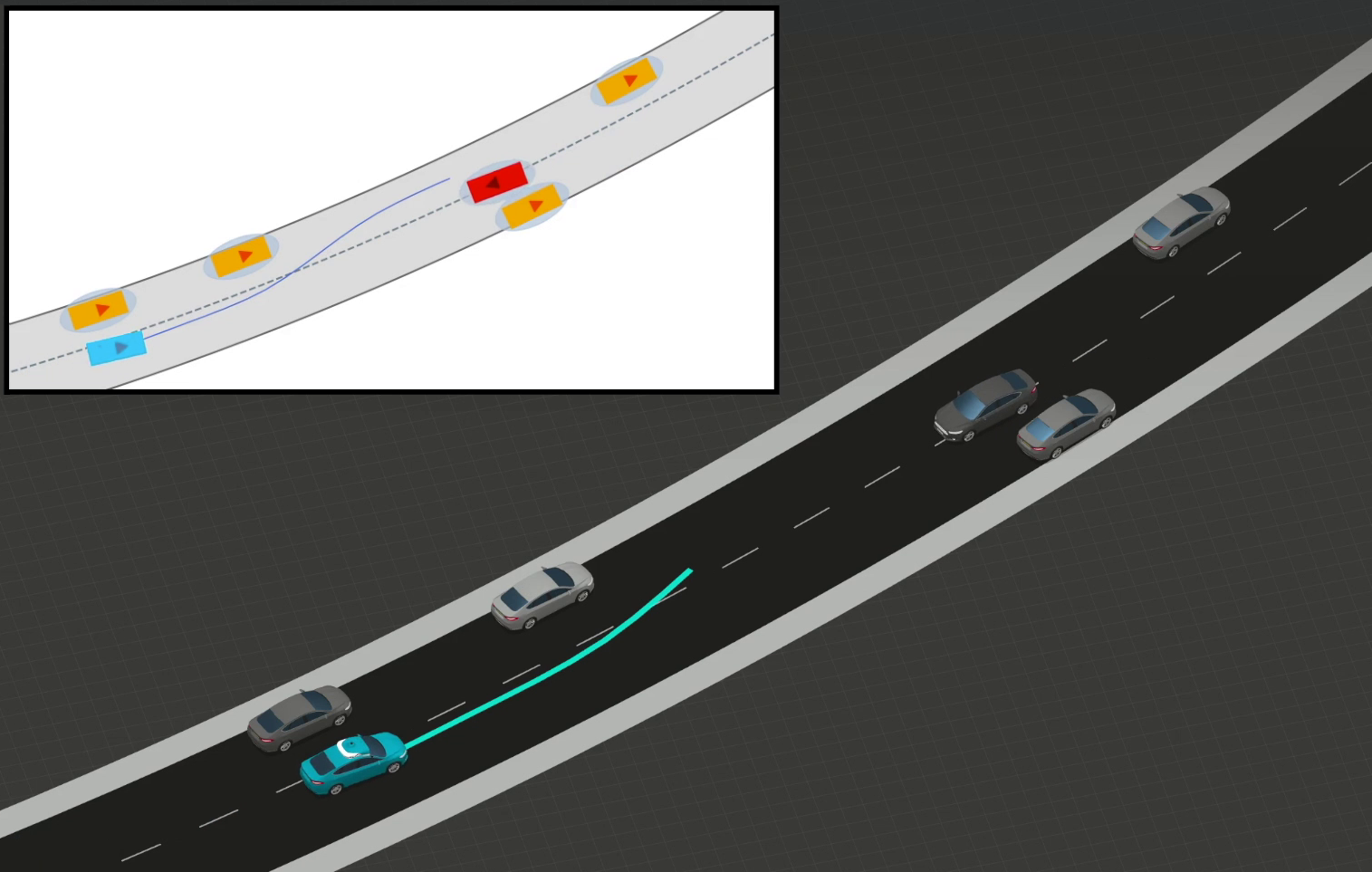}
        \label{fig:rd_sim}
    \end{subfigure}
    \vspace{-1.0em}
    \caption{\textit{Urban driving}: simulation view of our planner overtaking static vehicles while handling an oncoming vehicle, with the  planner's view in the inset. A video of this and other planning situations are available in the supplementary material.
    }
    \label{fig:view3D}
    \vspace{-1.0em}
\end{figure}

\IEEEPARstart{W}{ith} the rapid advancement of autonomous driving,
it is becoming increasingly clear that  guaranteeing safety across diverse driving scenarios is essential to the wider adoption of the technology~\cite{sw_kill, challenges}.
Performing safe, real-time planning in systems that must robustly achieve tight integration with scene perception and behavior prediction has continued to be an open challenge for the community~\cite{as2018aij, albrecht2020integrating}. This challenge is enhanced by the complex environments and decision-making that arises in an urban, residential driving setting~\cite{urban_driving}. Fig.~\ref{fig:view3D} shows an example of such environments.

An enticing approach to this problem, motivated by the ability to collect driving data from millions of miles driven by sensorized vehicles, is to exploit machine learning methods, such as in a deep imitation learning paradigm~\cite{wayve, nvidia}. While these methods have been shown to be successful in contained environments, it has been difficult to provide safety guarantees when policies are learned in this fashion, particularly with noisy training data and fault-susceptible real-time perception. Initial works aimed at providing robustness guarantees of neural networks, while encouraging, remain limited in scope in light of the scale of the models required in practical systems~\cite{safety_verification, parot}.

\begin{figure*}[t]
\centering
\includegraphics[width=0.9\textwidth]{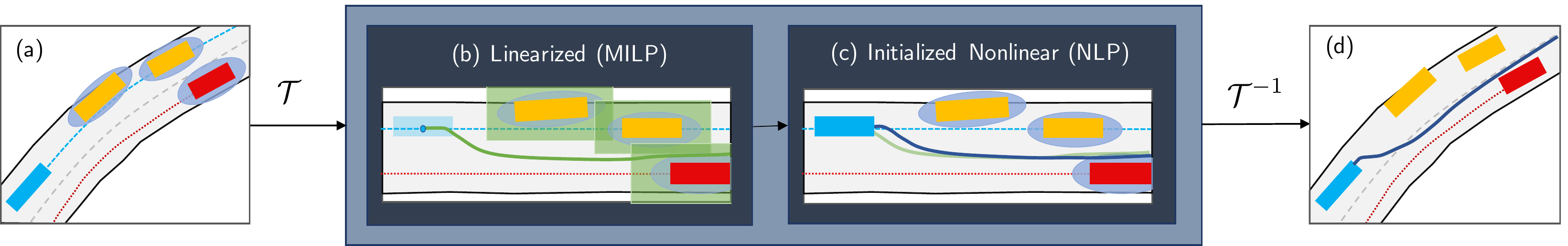}
\caption{\textit{A Two-Stage Optimization Planner Architecture}: (a) from an initial scene  - comprising driveable surface limits, static vehicles (yellow), moving vehicles with  predicted trajectories (red), and a reference path to follow (light blue) - a transform $\mathcal{T}$ yields the input to the planner in  the reference path-based coordinate frame. The MILP stage (b)  solves a linearized version of the planning problem, which initializes (c) the nonlinear, kinematically-feasible NLP stage. Then, $\mathcal{T}^{-1}$ transforms the output back to a trajectory in the world coordinate frame (d).}
\label{fig:overview}
\vspace{-1.0em}
\end{figure*}

The multi-dimensional requirements associated with planning in autonomous vehicles are inherently hierarchical in nature. The core concern of \textit{avoiding collisions} with other road users and obstacles, in the interest of passenger safety, is a hard requirement. Other secondary concerns -- such as continued progress towards a destination, comfort or power management -- imply softer requirements. Some planning methods, including many reinforcement learning formulations and some forms of unconstrained trajectory optimization, expect the hierarchy to be resolved implicitly through the design of the optimization objective~\cite{rl_plan, tebs}. However, such approaches tend to be brittle in enforcing the aforementioned safety-first hierarchy~\cite{safe_ai}.

A promising approach to such prioritization is seen in the formal methods literature, where logical specifications capture safety guarantees explicitly. Examples include optimizing the robustness signal of Signal Temporal Logic formulas in a receding horizon framework~\cite{stl, stl2}, and synthesis of policies that maximize the probability of satisfying objectives given as Linear Temporal Logic predicates~\cite{formal_rl}.
Despite their high quality output (and recent runtime improvements~\cite{fasterltl}) many of these tools lead to performance bottlenecks that are prohibitive in real-time settings~\cite{formal-survey, verifiable_ai}.

A more scalable technique to achieve prioritization is to adopt a constrained optimization approach, in which inviolable rules of the road~\cite{highway_code} are encoded as hard (in)equality constraints, while secondary objectives are satisfied in a soft manner as dictated by weights in a cost function~\cite{rick2019autonomous, liniger2015optimization, lam2012model, faulwasser2009model}.
In one approach, Schwarting \textit{et al.}~\cite{fast_nonlin} formulated planning as a nonlinear constrained optimization problem in a Model Predictive Control (MPC) scheme. While this formulation efficiently produces safe and smooth trajectories when it converges, the authors note some challenges with this approach including the uncertain convergence and/or locally-optimal solutions, even when using state-of-the-art solvers~\cite{fast_nonlin}. 
Due to the fact that nonlinear, non-convex constrained optimization methods are not guaranteed to converge even if feasible solutions exist~\cite{rick2019autonomous,local_optima_1,slow_convergence_1}, their safe deployment necessitates the implementation of an additional \textit{fallback mechanism} (e.g. emergency braking~\cite{fast_nonlin}), which might fall short in achieving the standard set by the safety hard constraints.

Our goal is to improve the convergence and solution quality of existing nonlinear constrained optimization methods, thus reducing the reliance on the potentially unsafe fallback mechanism and hence improving the safety of the overall system, while yielding lower-cost plans. We achieve this via an informed initialization step.
Specifically, we pose the planning problem in terms of a first stage modeled as a Mixed-Integer Linear Program (MILP), whose linearized, non-kinematically feasible output  initializes a Nonlinear Programming (NLP) problem, as shown in Fig.~\ref{fig:overview}.
The informed initialization offered by the first stage is an approximate solution that, theoretically, is globally $\varepsilon$-optimal\footnote{In practice, modern solver implementations  might introduce further limitations to the theoretical guarantee~\cite{MILP_optimal_2}.} up to a user-defined receding horizon~\cite{MILP_optimal_1}.
This initialization better enables the subsequent nonlinear, non-convex constrained optimization stage to converge to a safe, smooth and feasible trajectory of higher quality in terms of satisfying soft constraints. 
While this way of improving convergence and achieving higher quality solutions might compromise runtime efficiency,  it suits applications that can tolerate longer planning times in return for higher-quality plans~---~such as in applications of high-fidelity simulation or urban driving.

\textbf{Contributions} Our contributions are twofold:
\begin{enumerate}
    \item we formulate motion planning for urban driving in a novel two-stage constrained optimization framework, and
    \item we evaluate the framework thoroughly in a range of simulated complex urban driving situations to demonstrate empirically that our approach leads to a higher convergence rate and to lower cost solutions when compared with the alternative methods considered.
\end{enumerate}

\section{Preliminaries and Problem Statement}
\label{sec:preliminaries}

\subsection{Notation and Definitions}
\label{sec:notation}

We use the shorthand $k \equiv t_k = t_0 + k\Delta t$, where $t_0$ is current time and $\Delta t$ is the timestep. We assume the planning temporal horizon to be $\tau = N\Delta t$, defined over $N$ discrete steps. For any variable $r_j$ with $j \in \mathbb{N}^+$, we use the shorthand notation $r_{i:e} \equiv (r_i,...,r_e)$. Vectors are in boldface.

We refer to the vehicle we are planning for as the \textit{ego-vehicle} and consider its state at time $k$ to be given by $\mathbf{Z}_k = (X_k, Y_k, \Phi_k, V_k) \in \mathcal{W}_{\mathcal{Z}}$, where $(X_k, Y_k)$ is position, $\Phi_k$ is  heading, and $V_k$ is  speed, all in a global coordinate frame $\mathcal{W}$. The initial ego state at time 0 is the input $\mathbf{Z}_{0}$, while the output of planning is the vector of discrete ego states over the planning horizon, $\mathbf{Z}_{1:N} \in \mathcal{W}_{\mathcal{Z}}^{N}$.

Each other traffic participant (such as other vehicles, pedestrians and cyclists) is represented by a mean pose at time $k$, $\mathbf{O}_k^i = (X_k^i, Y_k^i, \Phi_k^i) \in \mathcal{W}_{\mathcal{O}}$, $i \in \{1,...,n\}$, and a Gaussian distribution for position uncertainty with mean $(X_k^i, Y_k^i)$ and covariance $\mathbf{\Gamma}_k^i$. The predicted poses and covariances of all traffic participants over the planning horizon $(\mathbf{O}_{0:N}^{1:n}$,   $\mathbf{\Gamma}_{0:N}^{1:n})$ are also inputs to the planning problem.

We define the planning objective to be continuous progress along a differentiable and bounded two-dimensional reference path $\pref$. The reference path, of length $|\pref|$, is parameterized by the longitudinal distance from its start $\lambda \in [0, |\pref|]$, with the points $(X^{\pref}(\lambda), Y^{\pref}(\lambda))$.

To simplify the planning problem, we transform the global coordinate frame $\mathcal{W}$ to a $\pref$-based representation $\mathcal{W}_r$ under the invertible transform $\mathcal{T}$, as presented in the next section. Fig.~\ref{fig:overview} illustrates the steps to transform a problem input $(\mathbf{Z}_0, \mathbf{O}_{0:N}^{1:n}, \mathbf{\Gamma}_{0:N}^{1:n}, \pref)$ to the desired planning output $\mathbf{Z}_{1:N}$ using $\mathcal{T}$, $\mathcal{T}^{-1}$ and our planner.

\subsection{Reference Path Representation}
\label{sec:ref_path_representation}

Given a reference path $\pref$, we can define its tangential and normal vectors in the global coordinate frame as:
\begin{equation}
\mathbf{t}_\lambda = \begingroup
\renewcommand*{\arraystretch}{1.5}
\begin{bmatrix} 
\frac{\partial X^{\pref}(\lambda)}{\partial \lambda} \\
\frac{\partial Y^{\pref}(\lambda)}{\partial \lambda}
\end{bmatrix}\endgroup \text{,}\quad \mathbf{n}_\lambda = \begingroup
\renewcommand*{\arraystretch}{1.5}\begin{bmatrix} 
\frac{-\partial Y^{\pref}(\lambda)}{\partial \lambda} \\
\frac{\partial X^{\pref}(\lambda)}{\partial \lambda}
\end{bmatrix}\endgroup
\label{tangents}
\end{equation}

\begin{figure}[t]
    \centering
    \includegraphics[width=0.45\textwidth]{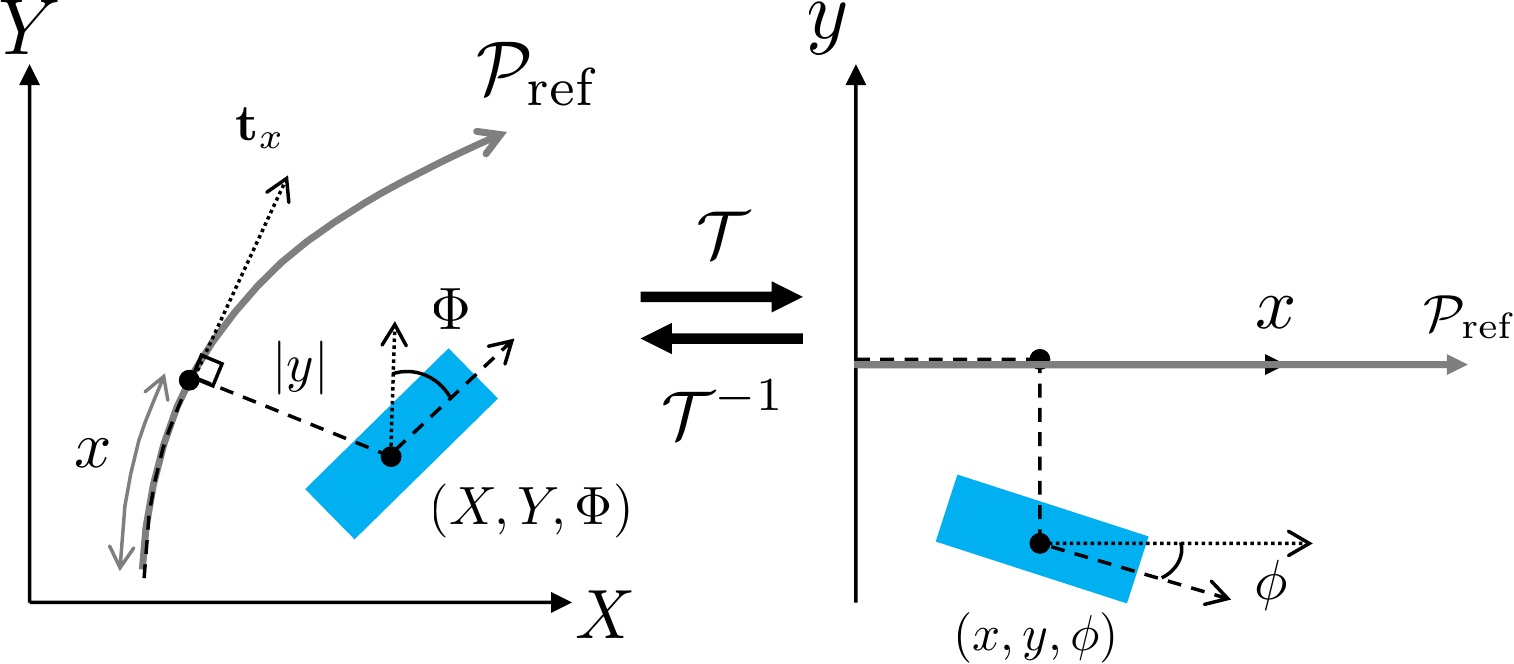}
    \caption{Visual representation of the pose transform of $\mathcal{T}$ from the global coordinate frame (left) to the reference path based coordinate frame (right), and the inverse transform $\mathcal{T}^{-1}$.}
    \label{fig:ref-path}
    \vspace{-1.0em}
\end{figure}

The invertible transform $\mathcal{T}$ operates on the input \textit{poses}, \textit{velocities} and \textit{covariance matrices} as follows.\footnote{The derivation of $\mathcal{T}^{-1}$ is similar to $\mathcal{T}$, and thus is omitted for brevity.}

\subsubsection{Pose transform} $\mathcal{T}$ maps the pose $(X, Y, \Phi)$ in the global coordinate frame $\mathcal{W}$ to a pose $(x, y, \phi)$ in the reference path frame $\mathcal{W}_r$, see Fig.~\ref{fig:ref-path}. 

\begin{itemize}
\item

$ x = \text{proj}_{\pref} \begin{bmatrix} 
X & Y
\end{bmatrix}
$
is the distance $\lambda$ from the beginning of $\pref$ to the projection of $\begin{bmatrix} 
X & Y
\end{bmatrix}$ into it, defined as:
\begin{equation*}
\label{eq:projection}
x = \underset{\lambda}{\text{argmin}} \quad (X - X^{\pref}(\lambda))^2 + (Y - Y^{\pref}(\lambda))^2.
\end{equation*}

Due to the nature of the optimization, no closed-form solution can be obtained for $x$. 

\item  $y = \frac{1}{||\mathbf{n}_x||}\,\mathbf{n}_x^\top\cdot \hat{\mathbf{y}}$, where $\mathbf{n}_x$ is the normal vector of the reference path at $\lambda = x$ as in \eqref{tangents}, and $\hat{\mathbf{y}} = \begin{bmatrix} 
X - X^{\pref}(x) \\
Y - Y^{\pref}(x)
\end{bmatrix}$.

\item  $\phi = \angle \mathbf{t}_x - \Phi$,
where:
\begin{equation}
\label{eq:arctan_tx}
\angle \mathbf{t}_x = \text{arctan} \left(\left.\frac{\partial Y^{\pref}(\lambda)}{\partial X^{\pref}(\lambda)}\right\rvert_{\lambda = x}\right).
\end{equation}
\end{itemize}

\subsubsection{Speed transform} $\mathcal{T}$ is a spatial transformation, so speeds are invariant: $v = \mathcal{T}(V) = V$. 

\subsubsection{Covariance transform} for a traffic participant with pose $\mathbf{O}$, covariance $\mathbf{\Gamma}$, and transformed pose $\mathcal{T}(\mathbf{O}) = \begin{bmatrix}
x & y & \phi
\end{bmatrix}^\top$, the transformed covariance matrix is given by:
\begin{equation}
\mathbf{\Sigma} = \mathcal{T}(\mathbf{\Gamma}) = R(\angle \mathbf{t}_{x} - \phi)\mathbf{\Gamma} R(\angle \mathbf{t}_{x} - \phi)^\top
\end{equation}
where $\angle \mathbf{t}_x$ is as defined in \eqref{eq:arctan_tx}, and $R(\varphi) \in SO(2)$ is the rotation matrix for angle $\varphi$.

\subsection{Problem Statement}
\label{sec:problem_statement}

The planning problem is defined in the reference path coordinate frame $\mathcal{W}_r$. 
The state of the ego-vehicle at time $k$ is given by $\mathbf{z}_k = (x_k, y_k, \phi_k, v_k) \in \mathcal{Z}$ where $(x_k, y_k) \in \mathbb{R}^2$ is the position, $\phi_k \in \mathbb{R}$ is its heading, and $v_k \in \mathbb{R}$ is its speed. The evolution of the state is given by a discrete general dynamical system:
\begin{equation}
\mathbf{z}_{k+1} = f_{\Delta t}(\mathbf{z}_k, \mathbf{u}_k),
\end{equation}
where $f_{\Delta t}$ is a discrete, nonlinear function parameterized by $\Delta t$, and $\mathbf{u}_k = (a_k, \delta_k) \in \mathcal{U}$ is the acceleration and steering angle controls applied at time $k$. We consider the ego-vehicle to be a rigid body occupying an area $S_e \subset \mathbb{R}^2$ relative to its center, and denote the area occupied by the ego-vehicle at state $\mathbf{z}_k$ by $\mathcal{S}(\mathbf{z}_k) \subset \mathbb{R}^2$.

For other traffic participants $i \in \{1,...,n\}$, pose at time $k$ is given by $\mathbf{o}_k^i = (x_k^i, y_k^i, \phi_k^i) \in \mathcal{O}$ and position covariance by $\mathbf{\Sigma}_k^i$. Following the definition from~\cite{fast_nonlin}, we denote the area each traffic participant occupies with probability higher than $p_{\epsilon}$ by $\mathcal{S}^i(\mathbf{o}_k^i, \mathbf{\Sigma}_k^i, p_{\epsilon}) \subset \mathbb{R}^2$.

We define the driveable surface area $\mathcal{B} \subset \mathbb{R}^2$ to be the area in which it is safe for the ego-vehicle to drive in the reference path coordinate frame, and the unsafe area $\mathcal{B}_{out} = \mathbb{R}^2 \setminus \mathcal{B}$. With a cost function $J(\mathbf{z}_{0:N}, \mathbf{u}_{0:N-1})$ defined over the ego-vehicle's positions and controls, we can now pose the planning problem.

\begin{problem}[Motion Synthesis]
\label{prob:nlp}
Given an initial ego state $\mathbf{z}_0$, and  trajectories of other traffic participants $(\mathbf{o}_{0:N}^{1:n} , \mathbf{\Sigma}_{0:N}^{1:n})$ over the horizon $N$, compute:
\begingroup\makeatletter\def\f@size{10}\check@mathfonts
$$
\begin{aligned}
& \underset{\mathbf{z}_{1:N}, \mathbf{u}_{0:N-1}}{\text{argmin}}
& & J(\mathbf{z}_{0:N}, \mathbf{u}_{0:N-1}) \\
& \text{\hspace{2em} s.t.} & & \forall k \in \{0,..., N\}: \\
& & & \mathbf{z}_{k+1} = f_{\Delta t}(\mathbf{z}_k, \mathbf{u}_k) \\
& & & \mathcal{S}(\mathbf{z}_k) \cap \mathcal{B}_{out} = \emptyset \\
& & & \mathcal{S}(\mathbf{z}_k) \cap \left[\bigcup_{i \in \{1,...,n\}} \mathcal{S}^i(\mathbf{o}_k^i, \mathbf{\Sigma}_k^i, p_{\epsilon})\right] = \emptyset\\
\end{aligned}
$$
\endgroup
\end{problem}

\section{Nonlinear Programming Formulation}
\label{sec:nlp}

In this section, we describe a specific solution method to Problem~\ref{prob:nlp}, posing it as an NLP problem comprising: (1) \textit{kinematic vehicle model} constraints on the model transitions and allowed controls; (2) \textit{driveable area collision avoidance} constraints; (3) \textit{traffic participants' collision avoidance} constraints; and (4) a \textit{multi-objective cost function} over soft constraints. We will discuss these in order.

\subsubsection{Vehicle Model}
\label{sec:nlp-kinematics}
We consider a discrete kinematic bicycle model based on the center of the vehicle. Modeling the ego-vehicle as a rectangle with an inter-axle distance $L$, we write:
\begin{equation}
\label{eq:nlp-kinematic}
\begin{bmatrix} 
x_{k+1} \\
y_{k+1} \\
\phi_{k+1} \\
v_{k+1}
\end{bmatrix} = \begin{bmatrix} 
x_{k} \\
y_{k} \\
\phi_{k} \\
v_{k}
\end{bmatrix} + 
\begin{bmatrix} 
v_k\cos(\phi_k + \delta_k) \\
v_k\sin(\phi_k + \delta_k) \\
\frac{2v_k}{L}\sin(\delta_k)\\
a_k
\end{bmatrix}\Delta t
\end{equation}
We additionally limit the maximum allowed steering $|\delta_k| \leq \delta_{\max}$ and acceleration $a_{\min} \leq a_k \leq a_{\max}$, as well as maximum jerk $|a_{k+1} - a_k| \leq \dot{a}_{\max} \Delta t$ and angular jerk $|\delta_{k+1} - \delta_k| \leq \dot{\delta}_{\max} \Delta t$, to stay within the operational domain and for passenger comfort\footnote{The use of the same positive and negative jerk magnitude mirrors that of previous works~\cite{fast_nonlin, wurts2018collision} and was observed to work in practice, however the framework can be trivially extended to use different jerk bounds.}. We also constrain speed, $0 \leq v_{\min} \leq v_k \leq v_{\max}$, to maintain forward motion below the set speed limit.

\subsubsection{Collision Avoidance with driveable surface boundaries}
\label{sec:nlp-collision}
The limits of the driveable surface can include both road limits and roadworks. The driveable surface constraint can be written as  $\mathcal{S}(\mathbf{z}_k) \cap \mathcal{B}_{out} = \emptyset$. %
To define an adequate representation of the area of the ego-vehicle $\mathcal{S}(\mathbf{z}_k)$ at state $\mathbf{z}_k$, we relax the definition of the boundary of the ego-vehicle to only considering its corners. For a rectangular vehicle of width $w$ and length $l$, the positions of the corners at $\mathbf{z}_k$ are:%
\begin{equation}
\mathbf{c}^\alpha(\mathbf{z}_k) = \begin{bmatrix} 
x^\alpha_k \\
y^\alpha_k
\end{bmatrix} = R(\phi_k) \left(\alpha^\top\circ\begin{bmatrix} 
w/2 \\
l/2
\end{bmatrix}\right) + \begin{bmatrix} 
x_k \\
y_k
\end{bmatrix}
\label{corners}
\end{equation}
where $\alpha \in \mathcal{A} = \{\begin{bmatrix}1 & 1\end{bmatrix}, \begin{bmatrix}-1 & 1\end{bmatrix}, \begin{bmatrix}-1 & -1\end{bmatrix}, \begin{bmatrix}1 & -1\end{bmatrix}\}$, $R(\phi_k) \in SO(2)$ is the rotation matrix for heading $\phi_k$, and $\circ$ is the element-wise product.

With this, the constraint can be reduced to:
\begin{equation}
\label{eq:nlp-col-avoidance-1}
\forall \alpha \in \mathcal{A}: \mathbf{c}^\alpha(\mathbf{z}_k) \in \mathcal{B}.
\end{equation}

We assume the driveable surface $\mathcal{B}$ is described by two continuous and smooth functions of $x$; the left border $b_l(x)$ and the right border $b_r(x)$, such that $\forall x: b_l(x) > b_r(x)$. These boundaries impose a constraint on $y$, the lateral path deviation at position $x$. Then, \eqref{eq:nlp-col-avoidance-1} can be reduced to the set of constraints 
\begin{equation}
\forall \alpha \in \mathcal{A}: b_l(x^\alpha_k) \geq y^\alpha_k \geq b_r(x^\alpha_k),
\end{equation}
with $\begin{bmatrix}x^\alpha_k  & y^\alpha_k\end{bmatrix}^\top$ defined as in \eqref{corners}.

\subsubsection{Collision Avoidance with other traffic participants}
We will refer to the traffic participants as vehicles in this section, but other participants (like cyclists and pedestrians) could be handled similarly.
We model the area of other vehicles  $\mathcal{S}^i(\mathbf{o}_k^i, \mathbf{\Sigma}_k^i, p_{\epsilon})$ in a similar fashion to~\cite{fast_nonlin}.
Assuming $a_{\text{shape}}$, $b_{\text{shape}}$ to be the parameters of an ellipse $\mathcal{L}$ that conservatively inscribes the vehicle's shape with $\mathbf{\Sigma}^i_k$  axis-aligned to the vehicle's axis, we write:
\begin{equation}
\label{eq:ellipse_def}
\mathcal{S}^i
\subset \mathcal{L}(a_{\mathbf{\Sigma}^i_k} + a_{\text{shape}}, b_{\mathbf{\Sigma}^i_k} + b_{\text{shape}}) = \mathcal{L}(a^i_k, b^i_k)
\end{equation}
where $\mathcal{L}(a^i_k, b^i_k)$ inscribes  vehicle $i$ at time $k$ up to an uncertainty $p_{\epsilon}$, see~\cite{fast_nonlin} for more details.
Thus, we write the constraints:
\begin{equation}
g^{i, \alpha}(\mathbf{z}_k) > 1, \;\;\forall \alpha \in \mathcal{A},
\end{equation}
\begingroup\makeatletter\def\f@size{7}\check@mathfonts
\begin{equation*}
\begin{aligned}
g^{i, \alpha}(\mathbf{z}_k) = \begin{bmatrix} 
x^\alpha_k - x^i_k \\
y^\alpha_k - y^i_k
\end{bmatrix}^\top R\left(\phi^i_k\right)^\top \begin{bmatrix} 
{(a_k^i)^{-2}} & 0 \\
0 & {(b_k^i)^{-2}}
\end{bmatrix} R\left(\phi^i_k\right) \begin{bmatrix} 
x^\alpha_k - x^i_k \\
y^\alpha_k - y^i_k
\end{bmatrix}
\end{aligned}
\end{equation*}
\endgroup

\begin{figure}[t]
    \centering
    \includegraphics[width=0.48\textwidth]{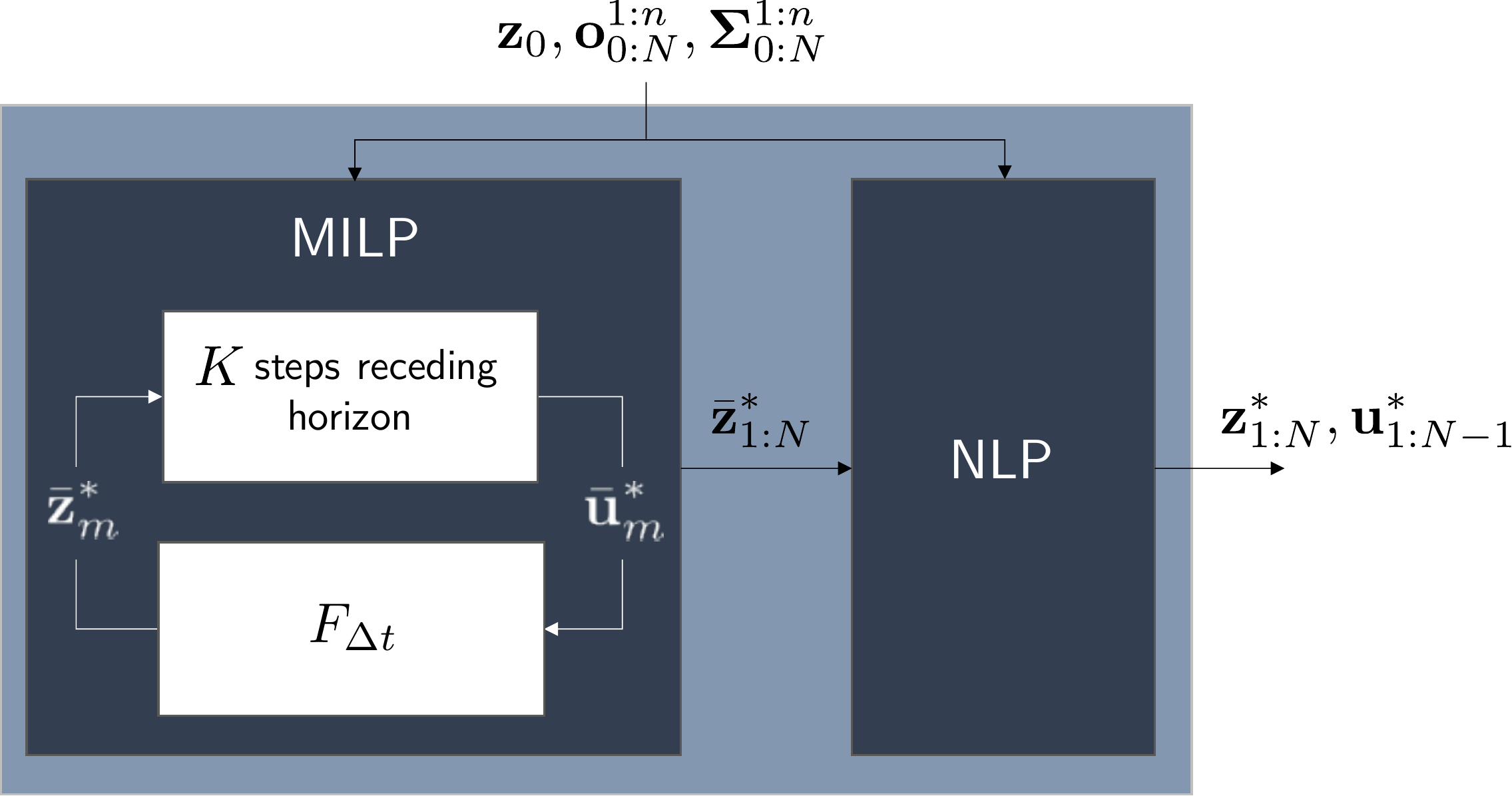}
    \caption{\textit{Detailed architecture}: the first stage (left) solves a receding horizon formulation of the MILP problem; the second stage (right) uses the solution of the first stage as initialization to solve the  full nonlinear, non-convex, constrained optimization problem.}
    \label{fig:cs-two-tier}
    \vspace{-0.8em}
\end{figure}

\subsubsection{Cost function}
\label{sec:nlp-cost}
A multi-objective cost function to be minimized is defined over the set of soft constraints $\mathcal{I}$ on the ego-vehicle's states and controls. For a soft constraint function %
$\theta_\iota(\mathbf{z}, \mathbf{u})$ and weight $\omega_\iota \in \mathbb{R}^+$ reflecting its relative importance, the cost function can be defined as:
\begingroup\makeatletter\def\f@size{10}\check@mathfonts
\begin{equation}
J(\mathbf{z}_{0:N}, \mathbf{u}_{0:N-1}) = \sum_{k=0}^{N}
\sum_{\iota \in \mathcal{I}} \omega_\iota \theta_\iota(\mathbf{z}_k, \mathbf{u}_k)
\end{equation}
\endgroup
Soft constraints target various objectives, including:
\begin{itemize}
	\item {\bf Progress} towards a longitudinal goal $x_g$: $\theta_x = (x - x_g)^2$, a target speed $v_g$: $\theta_v = (v - v_g)^2$, or minimizing lateral deviation from the reference path, $\theta_y = y^2$.
	\item {\bf Passenger comfort} defined as minimization of the norm of acceleration, $\theta_a = a^2$, and steering, $\theta_\delta = \delta^2$.
\end{itemize}

\subsection*{}
With the constraints and the cost function defined, we can now formulate the optimization  as a constrained nonlinear program.

\begin{problem}[Nonlinear Programming Problem]
\label{prob:nlp-specific}
Given an initial ego-vehicle state $\mathbf{z}_0$, trajectories of traffic participants $(\mathbf{o}_{0:N}^{1:n}, \mathbf{\Sigma}_{0:N}^{1:n})$ and soft constraints $\mathcal{I}$, compute:
\begingroup\makeatletter\def\f@size{10}\check@mathfonts
$$
\begin{aligned}
& \underset{\mathbf{z}_{1:N}, \mathbf{u}_{0:N-1}}{\text{argmin}}
& & J(\mathbf{z}_{0:N}, \mathbf{u}_{0:N-1}) \\
& \text{\hspace{2em} s.t.} & & \forall k \in \{0,..., N\}: \\
& & & \mathbf{z}_{k+1} = f_{\Delta t}(\mathbf{z}_k, \mathbf{u}_k) \\
& & & |\delta_k| \leq \delta_{\max}\\
& & & a_{\min} \leq a_k \leq a_{\max}\\
& & & |a_{k+1} - a_k| \leq \dot{a}_{\max}\Delta t\\
& & & |\delta_{k+1} - \delta_k| \leq \dot{\delta}_{\max}\Delta t\\
& & & v_{\min} \leq v_k \leq v_{\max}\\
& & & b_l(x^\alpha_k) \geq y_k^\alpha \geq b_r(x^\alpha_k), \alpha \in \mathcal{A}\\
& & & g^{i, \alpha}(\mathbf{z}_k) > 1, i \in \{1,...n\}, \alpha \in \mathcal{A}\\
\end{aligned}
$$
\endgroup
\end{problem}

Due to the nonlinearity of $J$, $f_{\Delta t}$, $b_l$, $b_r$ and $g^{i, \alpha}$, this is a nonlinear, non-convex, constrained optimization problem with equality and inequality constraints. While it is appealing to attempt to solve the problem directly or using a receding horizon formulation as in~\cite{fast_nonlin}, there are two major challenges:
\begin{itemize}
\item \textbf{Uncertain convergence}: solvers for this type of problems are generally slow for large instances, especially when initialization is not carefully considered~\cite{local_optima_1}. While advances in efficient primal-dual interior point solvers have mitigated this issue to a certain degree~\cite{ip_1,ip_2,ip_3}, the convergence to a solution is uncertain~\cite{fast_nonlin,local_optima_1,slow_convergence_1}.
\item \textbf{Local optima}: nonlinear constrained optimization solvers tend to be local in nature, finding  solutions that are close to the initial point and, possibly, far from the desired global optimum~\cite{fast_nonlin, MILP_optimal_1,local_optima_1}.
\end{itemize}

\section{Two-Stage Optimization}
\label{sec:two-step}

To mitigate the aforementioned issues of NLP, we propose the two-stage optimization framework presented in  Fig.~\ref{fig:cs-two-tier}. The main motivation behind the architecture is to %
improve the quality of the initial solution provided to the NLP solver, leading to faster and more reliable convergence overall. The informed initial solution is generated through a precursory optimization procedure.

In our two-stage framework, the first stage solves a \emph{linearized} version of Problem~\ref{prob:nlp-specific} formulated as a Mixed-Integer Linear Program (details in Sec.~\ref{sec:milp}) in a finite, receding horizon manner. Even though this only gives MILP optimality guarantees for each individual step of the receding horizon (see the discussion in  Sec.~\ref{sec:milp-optimality}), it does act as a proxy toward reaching the global optimum of the full linearized problem. The receding horizon process avoids the high runtime that would otherwise be incurred if the MILP problem is attempted with the full $N$ steps directly.

With the output of the MILP optimization taken as informed initialization, the second stage solves Problem~\ref{prob:nlp-specific}. If the  representations in the linearized and nonlinear problems are similar, this initialization should improve convergence, speed and the quality of the final solution.

\subsection{Mixed-Integer Linear Programming Formulation}
\label{sec:milp}

To reformulate Problem~\ref{prob:nlp-specific} as a MILP problem we (1)~consider a linear vehicle model with kinematic feasibility constraints; (2)~devise an approach to collision avoidance which maintains the mixed-integer linearity of the model; and (3)~adapt the soft constraints to a MILP cost function. %
We make use of the nonlinear operators $|\cdot|$ and $\max(\cdot)$ which we enforce through  auxiliary binary variables under the \textit{big-M} formulation~\cite{model_building, aerial_hybrid}.

\subsubsection{Linear Vehicle Model and Kinematic Feasibility}
\label{sec:milp-vehicle}

The nonlinear kinematic bicycle model presented in Sec.~\ref{sec:nlp-kinematics}  can be linearized around a point using a series expansion, but this approximation is only valid around the point and  yields higher errors as the distance to the point increases. To avoid this issue, we consider a linear, nonholonomic vehicle model instead.
We define the state of this linear vehicle model at time $k$ as $\overline{\mathbf{z}}_k = \begin{bmatrix}x_k & y_k & v_{k}^x & v_{k}^y\end{bmatrix} \in \mathcal{Z}^\mathcal{M}$, where
\begin{equation}
v_{k}^x = v_k\cos(\phi_k)\;,\;v_{k}^y = v_k\sin(\phi_k)
\end{equation}
 with controls $\overline{\mathbf{u}}_k = \begin{bmatrix}a_{k}^x & a_{k}^y\end{bmatrix} \in \mathcal{U}^\mathcal{M}$, and consider a linear vehicle dynamics model defined as:
\begin{equation}
\overline{\mathbf{z}}_{k+1} = F_{\Delta t}(\overline{\mathbf{z}}_k, \overline{\mathbf{u}}_k) = A_d\overline{\mathbf{z}}_k + B_d\overline{\mathbf{u}}_k
\end{equation}
where $F_{\Delta t}$ corresponds to a zero-order hold discretization of the continous state-space system:
\begin{equation}
\dot{\overline{\mathbf{z}}} = \begin{bmatrix} 
0 & 0 & 1 & 0 \\
0 & 0 & 0 & 1 \\
0 & 0 & 0 & 0 \\
0 & 0 & 0 & 0
\end{bmatrix}\overline{\mathbf{z}}^\top + \begin{bmatrix} 
0 & 0\\
0 & 0\\
1 & 0\\
0 & 1
\end{bmatrix}\overline{\mathbf{u}}^\top
\end{equation}

This nonholonomic model is quite simplistic when compared with the kinematic bicycle model in \eqref{eq:nlp-kinematic}, so in order to approximate kinematic feasibility\footnote{It should be noted that this model is not kinematically feasible with regard to the bicycle model, only approximately so.}, we add the constraint: 
\begin{equation}
v^x \geq \rho|v^y|
\end{equation}
for a constant $\rho \in \mathbb{R}^+$. Assuming forward motion, that is $\phi_k \in [-\frac{\pi}{2}, \frac{\pi}{2}]$, this constraint dictates that any movement in the $y$ direction requires motion in the $x$ direction as well.

We also enforce the same constraints as in the nonlinear model, in particular input bound constraints, $a^x_{\min} \leq a_{k}^x \leq a^x_{\max}$ and $a^y_{\min} \leq a_{k}^y \leq a^y_{\max}$; jerk  constraints, $|a_{k+1}^x - a_{k}^x| < \Delta a^x_{\max}\Delta t$ and $|a_{k+1}^y - a_{k}^y| < \Delta a^y_{\max} \Delta t$; and speed constraints $v^x_{\min} \leq v_{k}^x \leq v^x_{\max}$ and $v^y_{\min} \leq v_{k}^y \leq v^y_{\max}$, with $v^x_{\min} \geq 0$ to guarantee forward motion.

\subsubsection{Collision Avoidance}
\label{sec:milp-collision}

A key difference of the linear formulation to the nonlinear case with respect to collision avoidance constraints  is the lack of explicit representation of orientation in the linearized state $\overline{\mathbf{z}}$. A linear approximation of the  ego-vehicle's corners would thus induce large errors in the model. We instead consider the ego-vehicle to be a point mass  and augment the area it occupies to linearized approximations of the limits of the driveable surface and traffic participants.

For the driveable surface, we formulate the constraint:
\begin{equation}
d + b^\mathcal{M}_l(x_k) \geq y_k \geq b^\mathcal{M}_r(x_k) - d
\end{equation}
where $d: \mathcal{S}_e \to \mathbb{R}$ is a function of the shape of the ego-vehicle, and $b^\mathcal{M}_l(x)$ and $b^\mathcal{M}_r(x)$ are piecewise-linear functions for the left and right road boundaries such that $\forall x: b^\mathcal{M}_l(x) > b^\mathcal{M}_r(x)$. 
For a rectangular vehicle with width $w$ and length $l$, the most restrictive approximation of the shape of the ego-vehicle is given by $d=\sqrt{(w^2 + l^2)}/2$, limiting the driveable surface to $\mathcal{B} = \mathbb{R}^2 \setminus (\mathcal{B}_{out} \oplus \mathcal{S}_e)$ where $\oplus$ is the Minkowski-sum operator. For practical purposes, we consider $d = w/2$, which is exact when $\phi = 0$.

For traffic participants, we inscribe the ellipses $\mathcal{L}(a^i_k, b^i_k)$ defined in \eqref{eq:ellipse_def} with axis-aligned rectangles, augmented with the shape of the ego-vehicle. With $d^x$ and $d^y$ being functions of the size of the ego-vehicle with respect to its center in the $x$ and $y$ direction, respectively, we define  rectangles $\mathcal{R}^i_k$ with the limits:
\begin{equation}
\begin{aligned}
x^i_{k, \min} & = \left[\min_x \mathcal{L}(a^i_k, b^i_k)\right] - d^x \\
x^i_{k, \max} & = \left[\max_x \mathcal{L}(a^i_k, b^i_k)\right] + d^x \\
y^i_{k, \min} & = \left[\min_y \mathcal{L}(a^i_k, b^i_k)\right] - d^y \\
y^i_{k, \max} & = \left[\max_y \mathcal{L}(a^i_k, b^i_k)\right] + d^y.
\end{aligned}
\end{equation}
It should be noted that $x^i_{k, \min}, x^i_{k, \max}, y^i_{k, \min}, y^i_{k, \max}$ can be computed from $\mathcal{L}(a^i_k, b^i_k)$, $d_x$ and $d_y$ in closed form.
Then, the collision avoidance constraint is the logical implication:
\begin{equation}
(x^i_{k, \min} \leq x \leq x^i_{k, \max} \,\,\wedge\,\, y \geq y^i_{k, \min}) \,\,\Rightarrow\,\, y \geq y^i_{k, \max}
\end{equation}
which reads \textit{if the ego's position is aligned with a vehicle along $x$, then it must be outside the vehicle's borders in $y$}. Using the \textit{big-M} formulation~\cite{model_building, aerial_hybrid} with a sufficiently large $M \in \mathbb{R}^+$, we get the corresponding mixed-integer constraint:
\begin{multline}
\label{eq:milp-collision-avoidance}
y^i_{k, \max} - M\mu^i_k \leq y_k \;\text{where}\\
\mu^i_k = \max(x^i_{k, \min} - x_k, 0) + \max(x_k - x^i_{k, \max}, 0) \\+ \max(y^i_{k, \min} - y_k, 0)
\end{multline}

\subsubsection{Mixed-Integer Linear Cost Function}
\label{sec:milp-cost}

It is imperative that the MILP cost function be similar to the one from the nonlinear stage in order to minimize the gap between the optima of the two stages. With the receding horizon formulation, the cost over each state is also defined independently in time, as, for $k \in \{0,...,N\}$:
\begin{equation}
J_\mathcal{C}^{\mathcal{M}, k}(\overline{\mathbf{z}}, \overline{\mathbf{u}}) = \sum_{\iota \in \mathcal{C}} \Omega_\iota \Theta_\iota(\overline{\mathbf{z}}, \overline{\mathbf{u}}),
\end{equation}
for the set of soft constraints $\mathcal{C}$, where each constraint $\iota$ defines  a function to minimize, $\Theta_\iota(\overline{\mathbf{z}}, \overline{\mathbf{u}})$, and a weight $\Omega_\iota \in \mathbb{R}^+$ to determine  relative importance.%
 Similar to the nonlinear stage, soft constraints target various objectives:
\begin{itemize}
	\item \textbf{Progress} towards a longitudinal goal $x_g$, $\Theta_x = |x - x_g|$; a target speed $v_g$, $\Theta_v = |v - v_g|$; and  lateral deviation from the reference path, $\Theta_y = |y|$.
	\item \textbf{Passenger comfort}  as a minimization of the norm of lateral acceleration, $\Theta_a = |a_y|$.
\end{itemize}

\subsubsection{MILP Problem Definition}
\label{sec:milp-problem}

With constraints $\mathcal{C}$ and cost function $J^{\mathcal{M}, k}$, we can now formulate the planning problem as a $K$-step receding horizon, Mixed-Integer Linear Program. That is, we obtain $\overline{\mathbf{z}}_{1:N}^*$ by solving $N-K$ consecutive problems.

\begin{problem}[Receding Horizon MILP]
\label{prob:milp}
Given an initial ego-vehicle state $\overline{\mathbf{z}}_0$,  trajectories of other traffic participants $(\mathbf{o}_{0:N}^{1:n}, \mathbf{\Sigma}_{0:N}^{1:n})$, a set of soft constraints $\mathcal{C}$, compute for planning iteration $0 \leq m \leq N - K$ of the receding horizon:
\begingroup\makeatletter\def\f@size{10}\check@mathfonts
$$
\begin{aligned}
\label{eq:milp_problem}
& \underset{\overline{\mathbf{z}}_{m+1:m+K}, \overline{\mathbf{u}}_{m:m+K-1}}{\text{argmin}}
& & \sum_{k=m}^{m+K} J_\mathcal{C}^{\mathcal{M}, k}(\overline{\mathbf{z}}_k, \overline{\mathbf{u}}_k) \\
& \text{\hspace{4.2em} s.t.} & & \forall k \in \{m,..., m+K\}: \\
& & & \overline{\mathbf{z}}_{k+1} = F_{\Delta t}(\overline{\mathbf{z}}_k, \overline{\mathbf{u}}_k) \\
& & & v^x \geq \rho|v^y|\\
& & & a^x_{\min} \leq a_{k}^x \leq a^x_{\max}\\
& & & a^y_{\min} \leq a_{k}^y \leq a^y_{\max}\\
& & & |a_{k+1}^x - a_{k}^x| < \Delta a^x_{\max}\Delta t\\
& & & |a_{k+1}^y - a_{k}^y| < \Delta a^y_{\max} \Delta t\\
& & & v^x_{\min} \leq v_{k}^x \leq v^x_{\max}\\
& & & v^y_{\min} \leq v_{k}^y \leq v^y_{\max}\\
& & & d + b^\mathcal{M}_l(x_k) \geq y_k%
\geq b^\mathcal{M}_r(x_k) - d \\
& & & y^i_{k, \max} - M\mu^i_k \leq y_k, i \in \{1,...,n\}\\
\end{aligned}
$$
\endgroup
\end{problem}

\subsubsection{On the optimality of the MILP stage}
\label{sec:milp-optimality}

The optimal solution $\overline{\mathbf{z}}_{1:N}^*$ of Problem~\ref{prob:milp} is used as an initialization to warm-start Problem~\ref{prob:nlp-specific} in order to mitigate the convergence and local optimality challenges discussed above.

A solution to this MILP problem can be obtained using Branch and Bound, a divide and conquer algorithm first introduced and applied to MILP by Land and Doig~\cite{bb_1},  proven to return the global $\varepsilon$-optimal solution~\cite{MILP_optimal_1, MILP_optimal_2}. In practice, however, modern solvers (e.g. Gurobi or CPLEX) may fail to find that optimal solution due to rounding errors and built-in tolerances~\cite{MILP_optimal_2}. Moreover, the receding horizon formulation of Problem~\ref{prob:milp}, introduced for the sake of computational tractability, results in suboptimality~\cite{mpc_suboptimal_1,mpc_suboptimal_2}. Due to these factors, no strong theoretical guarantee can be given regarding the optimality of the MILP stage.

Our hypothesis, nevertheless, is that partial solutions that are close enough to the global optimum for each receding horizon iteration act as proxies towards a final, agglomerated solution that is, when compared with the global optimum, close enough to be useful in initializing the NLP stage in order to improve the quality of the final solution. Our experiments corroborate this hypothesis.

\begin{figure}[t]
    \centering
    \begin{subfigure}[b]{0.49\textwidth}
        \centering
        \includegraphics[width=\textwidth]{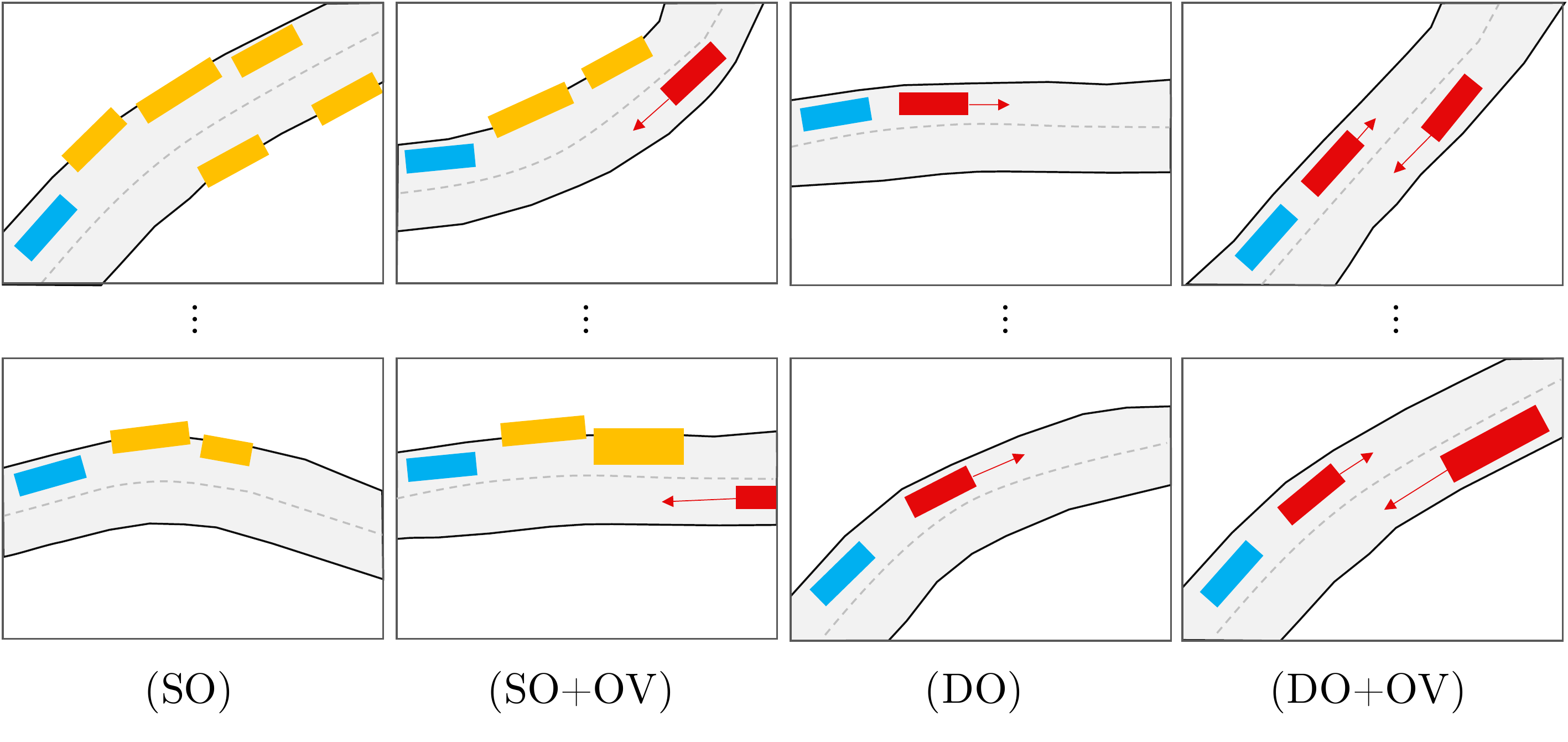}
    \end{subfigure}
    \vspace{-0.5em}
    \caption{\textit{Urban Driving Scenario Classes}: with static vehicles in yellow and dynamic vehicles in red, two examples each of: (\texttt{SO}) static overtaking of parked vehicles; (\texttt{SO+OV}) static overtaking with an oncoming vehicle in the other lane; (\texttt{DO}) dynamic overtaking of a slow moving vehicle; (\texttt{DO+OV}) dynamic overtaking of a slow moving vehicle with an oncoming vehicle in the other lane. Parameters, such as the map layout and number of vehicles in an example, are randomly generated.}
    \label{fig:scenarios}
    \vspace{-1.0em}
\end{figure}

\begin{figure*}[t]
    \centering

    \begin{subfigure}[b]{0.255\textwidth}
        \centering
        \includegraphics[height=2.0cm]{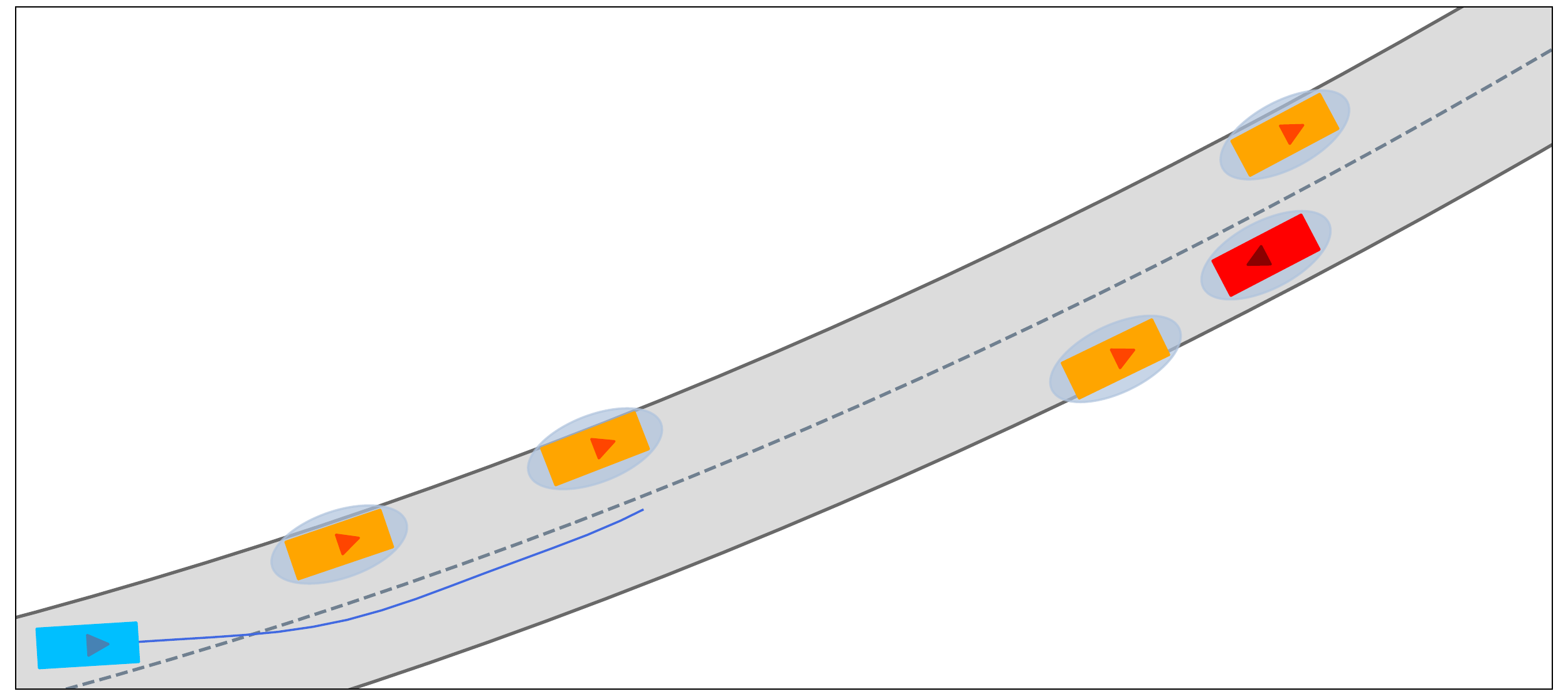}
        \caption{$t=0.0s$}
        \label{fig:qualitative_0_0}
    \end{subfigure}
    \begin{subfigure}[b]{0.24\textwidth}
        \centering
        \includegraphics[height=2.0cm]{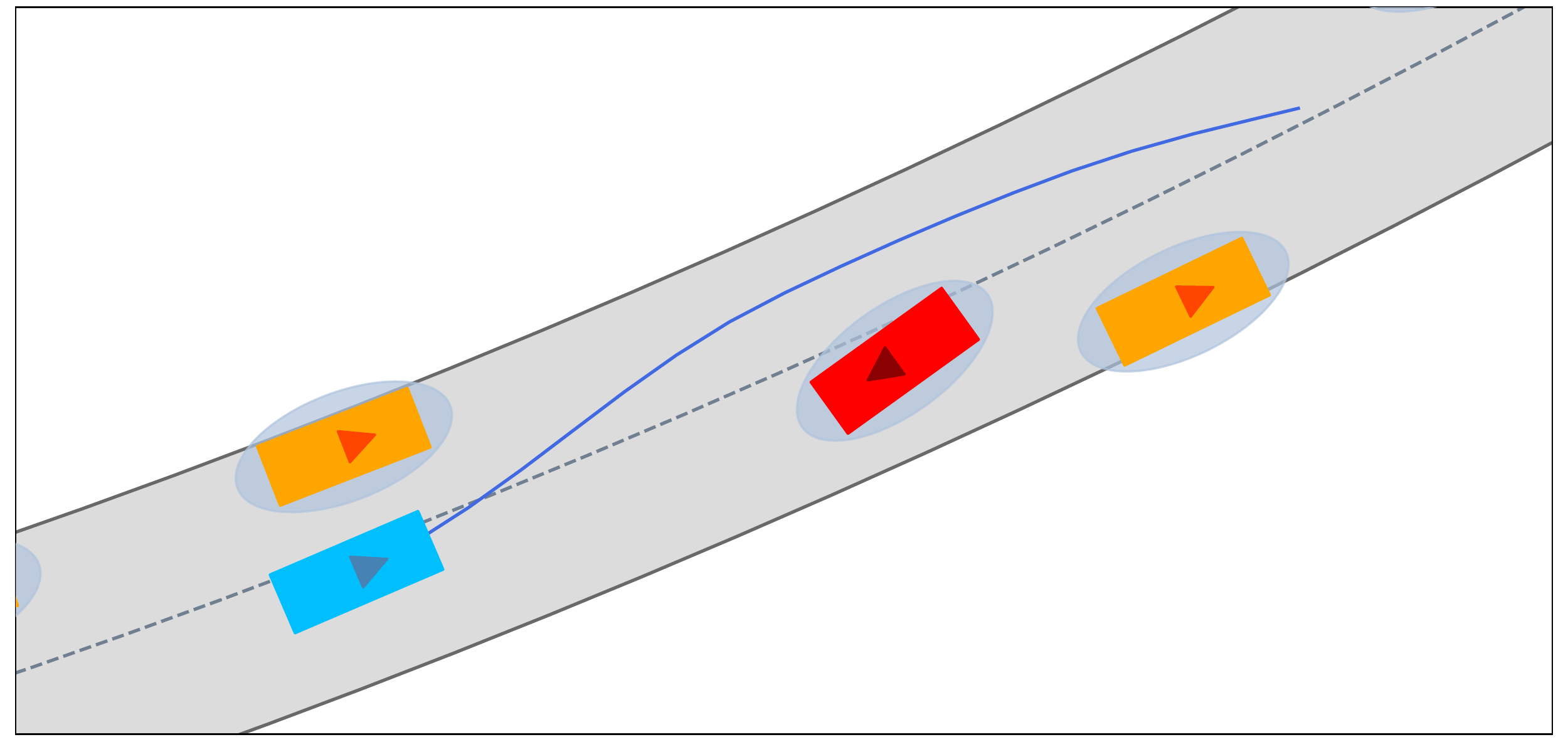}
        \caption{$t=3.2s$}
        \label{fig:qualitative_3_2}
    \end{subfigure}
    \begin{subfigure}[b]{0.23\textwidth}
        \centering
        \includegraphics[height=2.0cm]{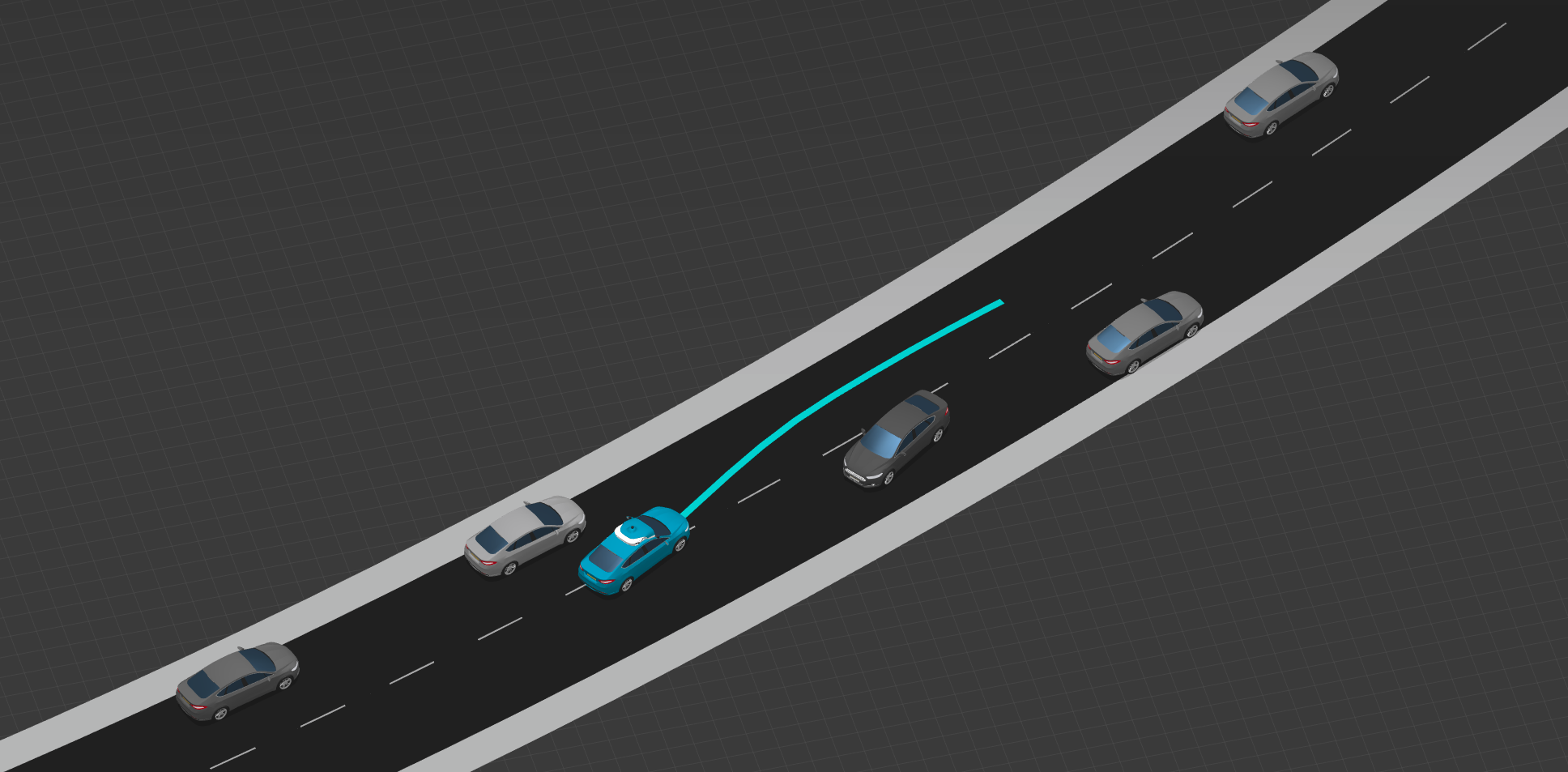}
        \caption{Simulator view ($t=3.6s$)}
        \label{fig:sim_view}
    \end{subfigure}
    \begin{subfigure}[b]{0.24\textwidth}
        \centering
        \includegraphics[height=2.0cm]{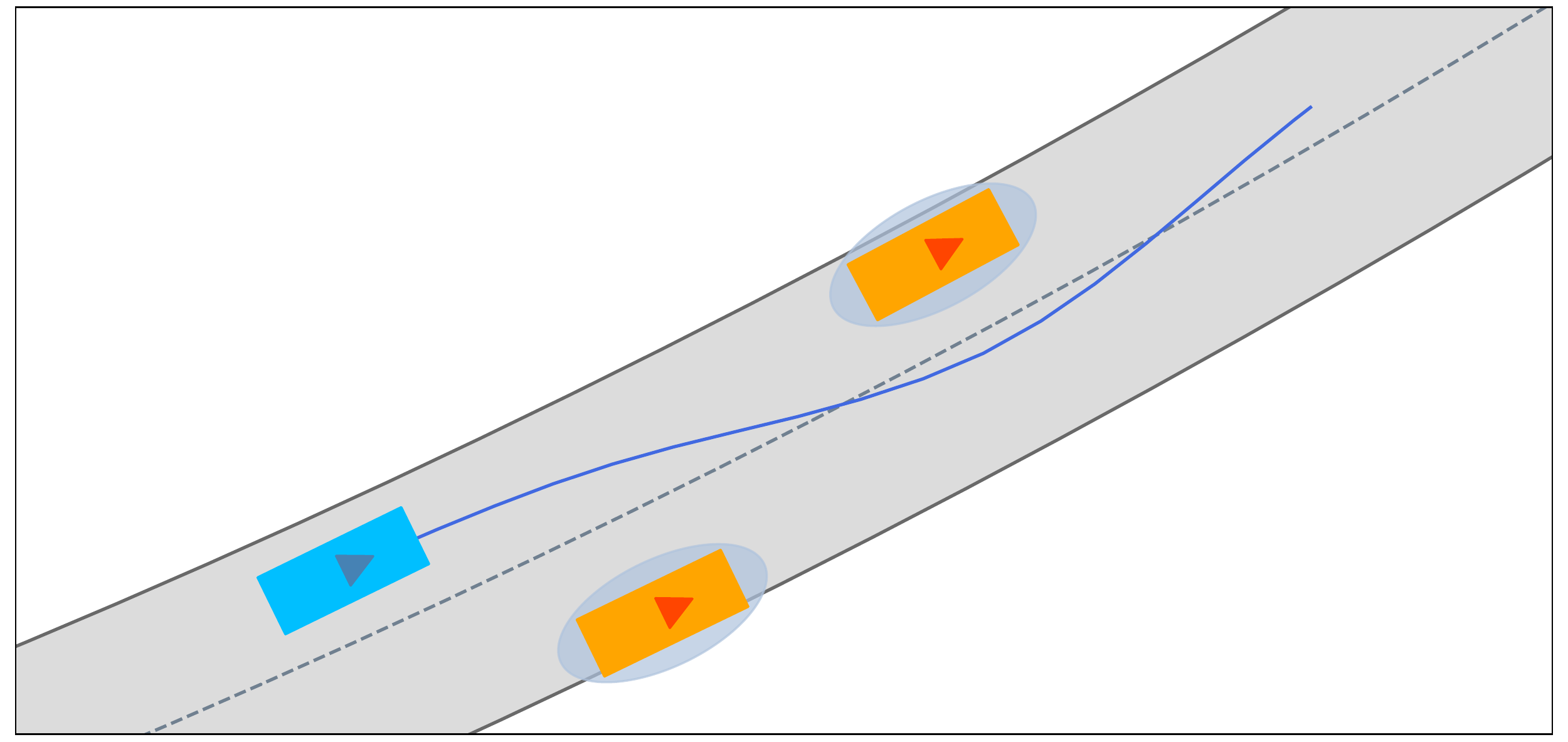}
        \caption{$t=5.2s$}
        \label{fig:qualitative_6_07}
    \end{subfigure}
    \vspace{-0.3em}
    \caption{Residential driving example: (a), (b) and (d) showing the planner's view of a residential driving problem, with the ego-vehicle (blue), static obstacles (orange), oncoming vehicle (red) and ego's plan (dark blue trace) at different times $t$; (c) shows our simulator's rendering of the situation at $t=3.6s$.
    }
    \label{fig:residential_three_plots}
\end{figure*}

\section{Experimental Results}
\label{sec:results}

\begin{table*}[!t]
\centering
\bgroup
\def\arraystretch{1.4}
\begin{tabular}{ll|c|cc|ccc}
& & \multirow{2}{*}{(a) Converged (\%)} & \multicolumn{2}{c|}{(b) Per instance comparison} & \multicolumn{3}{c}{(c) Overall runtime} \\ \cline{4-8}
& & & $\Delta_{\textsc{Ours}}^{\text{NLP}}$Cost (\%)  & $\Delta_{\textsc{Ours}}^{\text{NLP}}$ Runtime (\%) & Init (s)     & NLP (s) & Total (s) \\ \Xhline{2\arrayrulewidth}
\multirow{4}{*}{\textbf{Heuristics}} & \textsc{Zeros}  & $95.95$     & $11.62$     & $85.36$  & $0.0$       & $0.71\stddev{0.56}$       & $0.71\stddev{0.56}$ \\ 
& \textsc{Ct. Dec}  & $93.61$     & $9.64$     & $21.47$  & $0.0$       & $0.44\stddev{0.37}$       & $\mathbf{0.44}\stddev{0.37}$ \\ 
& \textsc{Ct. Vel}  & $63.88$     & $4.11$     & $37.97$  & $0.0$       & $0.63\stddev{0.89}$       & $0.63\stddev{0.89}$ \\
& \textsc{Ct. Acc}  & $39.89$     & $-0.65$     & $29.73$  & $0.0$       & $0.68\stddev{0.64}$       & $0.68\stddev{0.64}$ \\ \hline
\multirow{4}{*}{\textbf{MILP-based}} & \textsc{NoCol + NoVel} & $37.49$ & $0.27$ & $37.66$ & $0.14\stddev{0.08}$       & $0.70\stddev{0.66}$       & $0.84\stddev{0.67}$ \\
& \textsc{NoCol} & $42.08$ & $-1.18$ & $43.33$ & $0.17\stddev{0.08}$       & $0.65\stddev{0.48}$       & $0.83\stddev{0.49}$ \\
& \textsc{NoVel} & $93.22$ & $3.60$ & $15.70$ & $0.71\stddev{0.35}$       & $0.42\stddev{0.46}$       & $1.14\stddev{0.66}$ \\
& \cellcolor{lightblue}\textsc{Ours}   & \cellcolor{lightblue}$\mathbf{97.76}$     & \cellcolor{lightblue}\textemdash      & \cellcolor{lightblue}\textemdash  & \cellcolor{lightblue}$0.55\stddev{0.37}$       & \cellcolor{lightblue}$\mathbf{0.38}\stddev{0.42}$       & \cellcolor{lightblue}$0.93\stddev{0.60}$ \\ \Xhline{2\arrayrulewidth}
\end{tabular}
\egroup
\caption{\textit{Initialization Ablation}: (a) percentage of problems the NLP stage solves when initialized with the different methods; (b) average percent change in NLP cost and runtime in examples solved by our method and the alternative initialization (positive and higher is worse for the alternative method, better for our method); (c) running time per initialization method, and corresponding NLP stage and total runtime, in the respective subset of examples solved in the dataset by the method (mean $\pm$ standard deviation).
}
\label{tab:init-comp-all}
\vspace{-0.8em}
\end{table*}

In the following subsections, we show that:
\begin{enumerate}
    \item for the NLP stage, our MILP formulation provides a \textit{better inititalization} when compared with simpler heuristics and ablations of Problem~\ref{prob:milp}, leading to higher convergence (i.e., percentage of solved cases), faster NLP solving times, and lower-cost solutions, when compared with the NLP stage initialized by those methods (Sec.~\ref{sssec:results-2}); and
    \item our two-stage method \emph{outperforms}, in metrics of progress and passenger comfort, an alternative approach based on a Nonlinear Model Predictive Control (NMPC) scheme similar to~\cite{fast_nonlin} (Sec.~\ref{sssec:results-3}).
\end{enumerate}

In our experiments, we used state-of-the-art off-the-shelf solvers for the two types of optimization problems, implementing the first stage (MILP; Problem~\ref{prob:milp}) using Gurobi~\cite{gurobi} and the second stage (NLP; Problem~\ref{prob:nlp-specific}) using IPOPT~\cite{ipopt}. Both solvers have a timeout of $25s$, after which the optimization is stopped. We use $N=40$ and $\Delta t=0.2s$ for an output trajectory horizon of $8$ seconds. Other parameters of the two problems are listed in Appendix~\ref{sec:appendix-params}.

\subsection{Simulation Environment}

Without loss of generality, we assumed left-hand traffic where drivers are expected to be on the left-hand side of the road. We use a proprietary simulator to run the experiments presented in this paper. It uses a fixed simulation frequency of 100Hz, and all agents are simulated using a kinematic bicycle model as described in Eq.~\ref{eq:nlp-kinematic}. Perfect localization and full state observability at simulation time are assumed as sensor data for all agents. Agents also have access to a high definition map with road and lane boundary information, as well as markings, all represented as smooth piecewise linear functions. All non-ego agents implement an Intelligent Driver Model\cite{idm}-based decision-making process, allowing them to follow a lane, and perform adaptive cruise control~\cite{acc} assuming a forward vehicle as well as basic overtaking maneuvers.

The ego-vehicle has access to a route planner, which given a goal yields a reference path in the world coordinate frame, and a deterministic constant velocity and lane following predictor for other agents\footnote{While this is the case for the simulation environment in our experiments, we note that our method can include uncertainty-aware predictions following the modeling from~\cite{fast_nonlin}.}. It plans synchronously at a frequency of 1Hz.

\subsection{Urban Driving Scenario Classes}
\label{ssec:scenarios-in-eval}

Fig.~\ref{fig:scenarios} shows examples of the four classes of urban, two-lane driving scenarios we consider for the quantitative experiments in subsections~\ref{sssec:results-2} and~\ref{sssec:results-3}. Each of the classes presents a set of planning challenges associated with residential driving, among which: smooth overtaking of parked vehicles, negotiating an oncoming vehicle in the other lane, and the decision to overtake a slow-moving front vehicle.

Fig.~\ref{fig:residential_three_plots} shows the solution to an example in which the ego-vehicle overtakes vehicles parked at the sides of a two-lane road while taking into account an oncoming vehicle that also has to overtake static vehicles on the other lane. These planning challenges are reflected in many driving situations. For example, we show in Appendix~\ref{sec:appendix-junction} another example of urban driving that requires a similar level of negotiation with other vehicles: an unprotected junction right turn. Videos of both situations are available in the supplementary material.

To showcase robustness, we perform our analysis over a procedurally generated dataset obtained by randomizing over relevant parameters in each class (e.g., lane widths, initial speeds or number of vehicles). More details about the procedural generation process are presented in Appendix~\ref{sec:appendix-gen}.

\subsection{NLP Initialization Ablation} 
\label{sssec:results-2}
We considered four NLP stage heuristic initializations as alternatives to our MILP stage:
\begin{itemize}
    \item \textsc{Zeros}, in which all states and controls are initialized to zero;
    \item \textsc{Ct. Vel}, in which  the ego-vehicle is assumed to maintain a constant velocity throughout the solution;
    \item \textsc{Ct. Acc}, in which  the ego-vehicle maintains a constant acceleration of $1ms^{-2}$ until the maximum speed is reached; and
    \item \textsc{Ct. Dec}, in which  the ego-vehicle maintains a constant negative acceleration of $-1ms^{-2}$ until it stops.
\end{itemize}

For a more thorough analysis, we also considered three ablations of the MILP formulation presented in Problem~\ref{prob:milp}:
\begin{itemize}
    \item \textsc{NoCol + NoVel}, a similar initialization to the MILP formulation of Problem~\ref{prob:milp}, excluding collision avoidance, speed limit and  target speed constraints;
    \item \textsc{NoCol}, a similar initialization to the MILP formulation of Problem~\ref{prob:milp}, excluding the collision avoidance constraints with other road users;
    \item \textsc{NoVel}, a similar initialization to the MILP formulation of Problem~\ref{prob:milp}, excluding the speed limit and  target speed constraints.    
\end{itemize}

\begin{figure}[t]
    \centering
    \begin{subfigure}{0.48\textwidth}
        \centering
        \includegraphics[width=\textwidth]{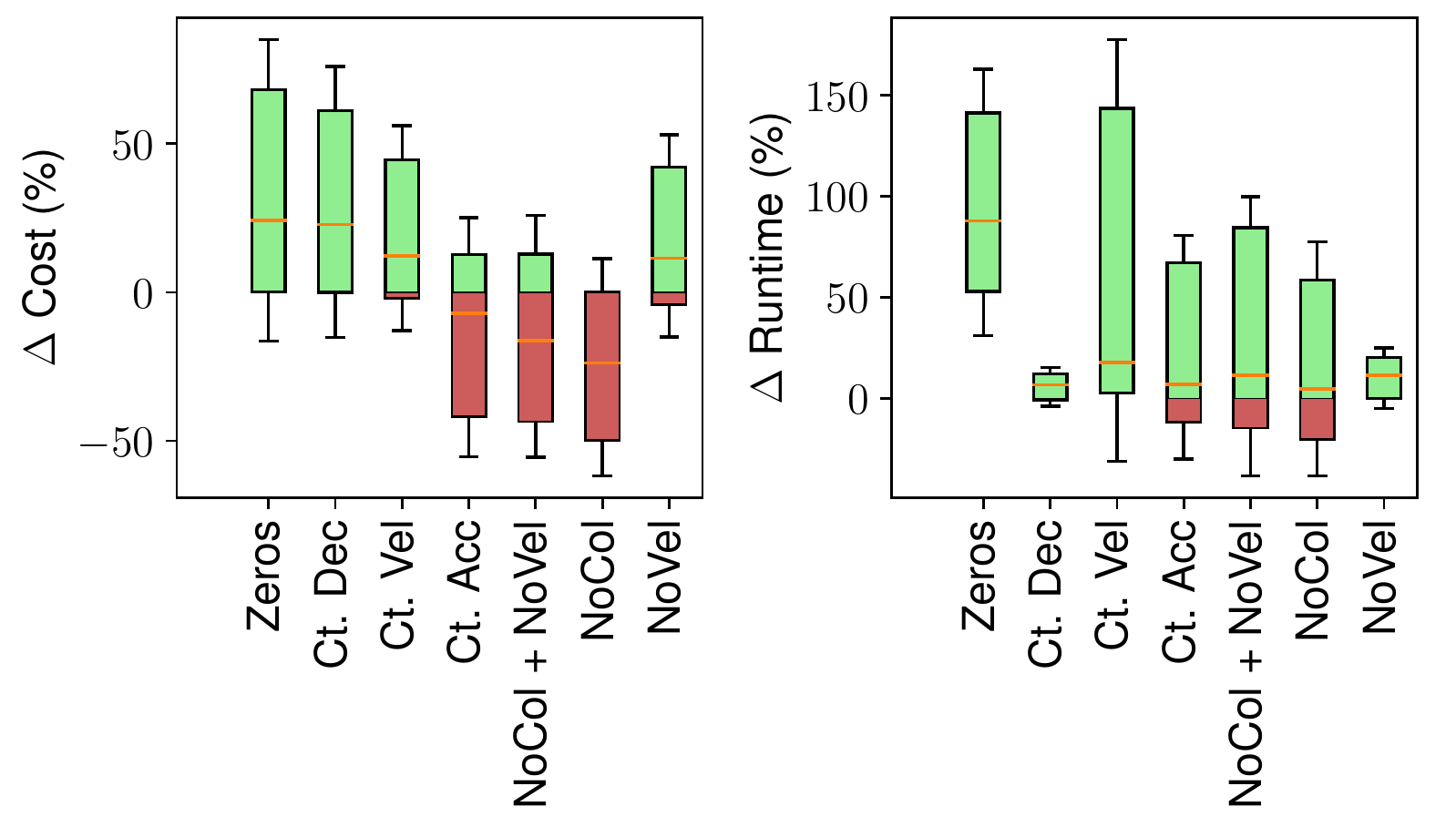}
    \end{subfigure}
    \vspace{-0.5em}
    \caption{\textit{Initialization Ablation}: box plots of the percent change in NLP cost (left) and runtime (right) for each method when compared to ours, in examples where both converged but to different optima. Other methods having percent change values larger than zero, $\Delta > 0$ (indicated by the green color), means better performance by our method.
    }
    \label{fig:init_boxplots}
    \vspace{-0.7em}
\end{figure}

\begin{figure}[t]
    \centering
    \begin{subfigure}[b]{0.45\textwidth}
        \centering
        \includegraphics[width=\textwidth]{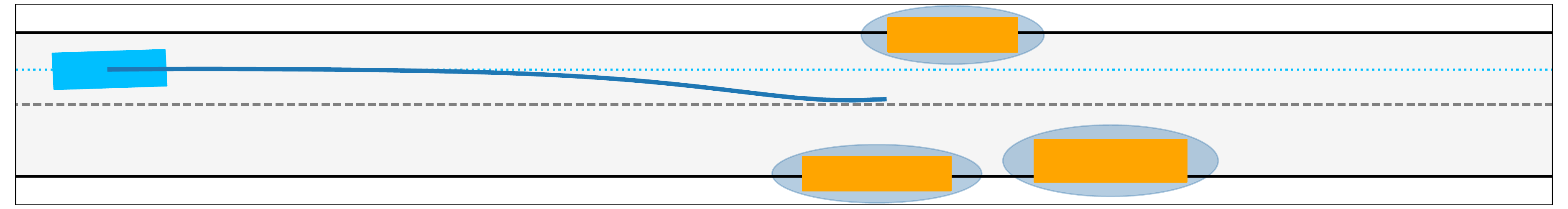}
        \caption{\textsc{Zeros}}
        \label{fig:init_zeros}
    \end{subfigure}
    \\\vspace{0.5em}
    \begin{subfigure}[b]{0.45\textwidth}
        \centering
        \includegraphics[width=\textwidth]{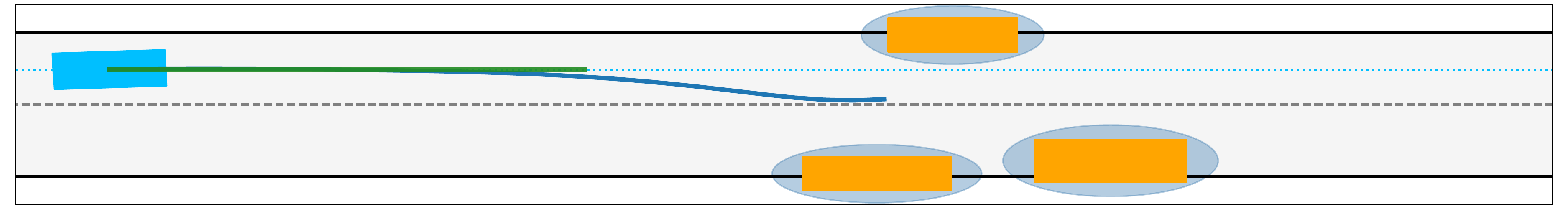}
        \caption{\textsc{Ct. Vel}}
        \label{fig:init_ct_vel}
    \end{subfigure}
    \\\vspace{0.5em}
    \begin{subfigure}[b]{0.45\textwidth}
        \centering
        \includegraphics[width=\textwidth]{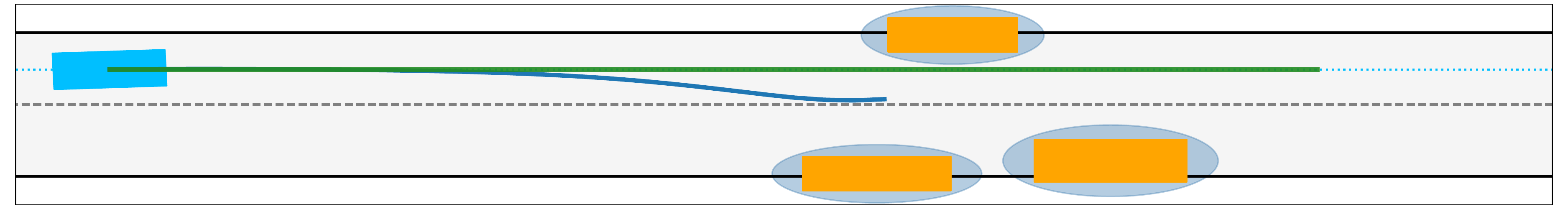}
        \caption{\textsc{Ct. Acc}}
        \label{fig:init_ct_acc}
    \end{subfigure}
    \\\vspace{0.5em}
    \begin{subfigure}[b]{0.45\textwidth}
        \centering
        \includegraphics[width=\textwidth]{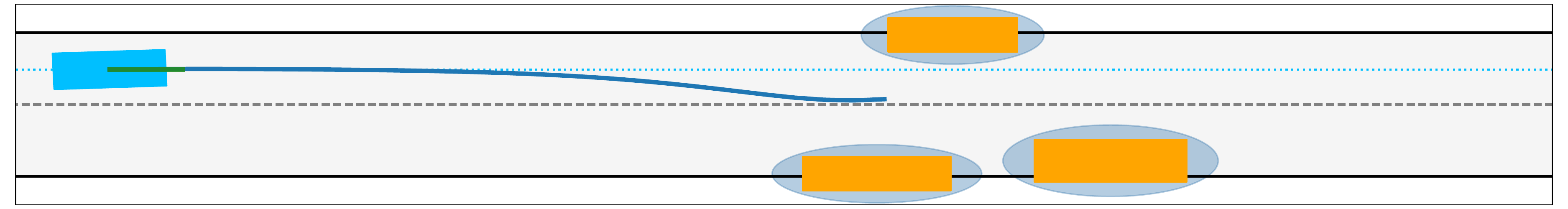}
        \caption{\textsc{Ct. Dec}}
        \label{fig:init_ct_dec}
    \end{subfigure}
    \\\vspace{0.5em}
    \begin{subfigure}[b]{0.45\textwidth}
        \centering
        \includegraphics[width=\textwidth]{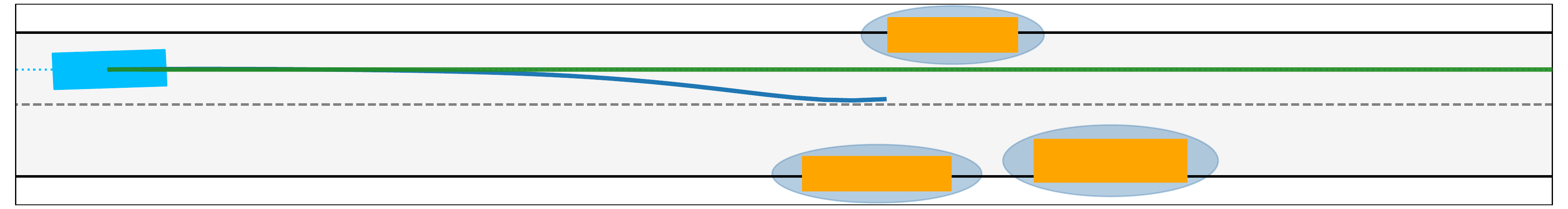}
        \caption{\textsc{NoCol + NoVel}}
        \label{fig:init_no_coll_no_vel}
    \end{subfigure}
    \\\vspace{0.5em}
    \begin{subfigure}[b]{0.45\textwidth}
        \centering
        \includegraphics[width=\textwidth]{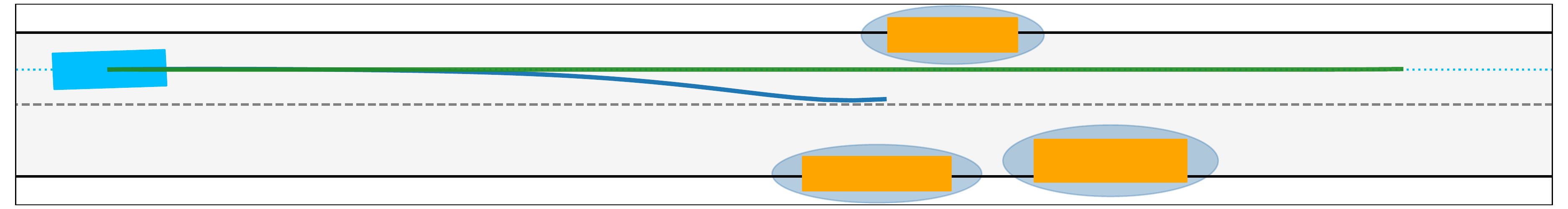}
        \caption{\textsc{NoCol}}
        \label{fig:init_no_coll}
    \end{subfigure}
    \\\vspace{0.5em}
    \begin{subfigure}[b]{0.45\textwidth}
        \centering
        \includegraphics[width=\textwidth]{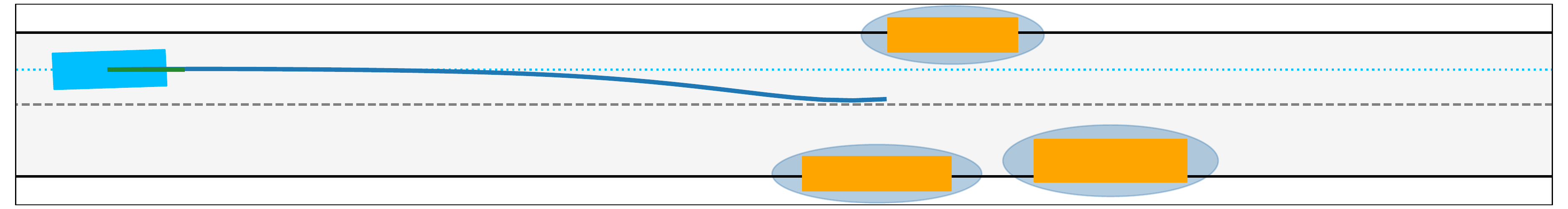}
        \caption{\textsc{NoVel}}
        \label{fig:init_no_vel}
    \end{subfigure}
    \\\vspace{0.5em}
    \begin{subfigure}[b]{0.45\textwidth}
        \centering
        \includegraphics[width=\textwidth]{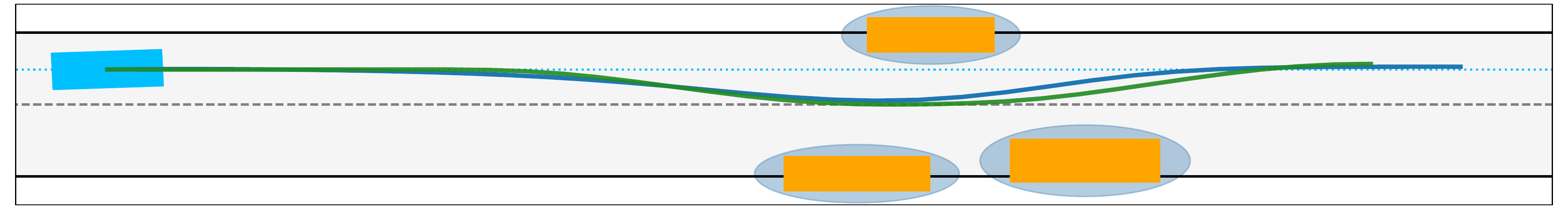}
        \caption{\textsc{Ours (MILP)}}
        \label{fig:init_ours}
    \end{subfigure}
    \\\vspace{0.5em}
    \caption{\textit{Initialization Ablation}: a qualitative example in the reference path coordinate frame of static overtaking from our dataset. The subfigures show the NLP output when initialized with (a)-(d) simple heuristics, (e)-(g) ablations of our MILP stage with subsets of constraints, and (h) our full MILP stage. The initialization trajectory is shown in green and the NLP stage output in dark blue.}
    \label{fig:ouput_diff_inits}
    \vspace{-0.8em}
\end{figure}

We generated a dataset of 4000 examples, 1000 examples per driving scenario class, and solved them using our method as well as with the NLP solver initialized by each of the alternative methods. All the NLP problems optimized the same constraints and cost function.

\begin{figure*}[t!]
    \centering
    \includegraphics[width=\textwidth]{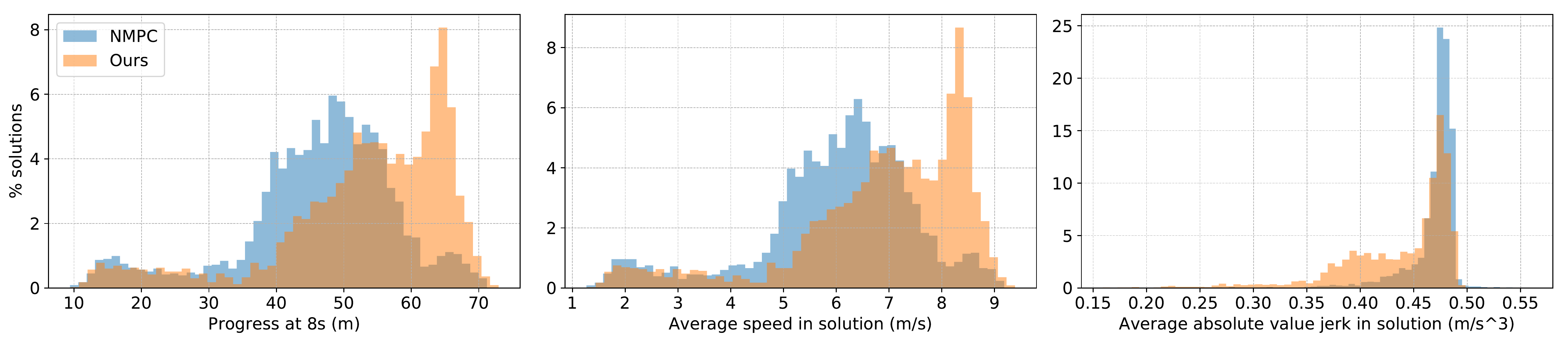}
    \caption{\textit{NMPC Comparison}: distributions of metric values for the subset of examples solved by both the NMPC (in blue) and our method (in orange). The metrics are: progress in meters after $8s$ (higher is better); average speed in an example (closer to the target of $8m/s$ is better); and average absolute value of jerk in an example (lower is better).}
    \label{fig:nmpc_comparison_distributions}
    \vspace{-0.8em}
\end{figure*}

From Table~\ref{tab:init-comp-all}~(a) we observe our MILP initialization leads to a higher percentage of solved examples than any other method considered. Table~\ref{tab:init-comp-all}~(b) shows a case-by-case comparison between our MILP initialization and the other methods in terms of the average final NLP cost and runtime for the subset of examples solved by both, 
showing that our MILP initialization achieves better NLP runtimes compared to all the considered alternatives and better costs in most of the cases. To analyze that further, in Fig.~\ref{fig:init_boxplots} we present boxplots of the distribution of the per instance change in NLP cost and runtime of a method compared to ours in examples where both converged to \textit{different} optima.
In most cases, our MILP stage leads to NLP solutions that outperform the alternatives in terms of cost, and that converge faster in the second stage. While \textsc{Ct. Acc}, \textsc{NoCol + NoVel} and \textsc{NoCol} appear to find better solutions in terms of cost in a non-negligible percentage of the considered examples in Fig.~\ref{fig:init_boxplots}, these approaches only converge to a valid solution in less than $43\%$ of the dataset examples as Table~\ref{tab:init-comp-all}~(a) shows and they are slower in general.
In Fig.~\ref{fig:ouput_diff_inits}, we show a qualitative example from the dataset in which all initializations are successful and our MILP initialization outperforms the alternative heuristics in terms of NLP cost.

Table~\ref{tab:init-comp-all}~(c) presents the running times of the initialization stage and the total runtime for the different methods we consider. Heuristic initializations have a negligible initialization time due to their simple closed-form formulations. %
Our MILP initialization leads to a lower NLP runtime than all other heuristic and ablation alternatives, indicating a higher-quality starting point that leads to quicker convergence. However, it does incur a higher total runtime than most due to the solving time in the first stage.

\subsection{NMPC Baseline Comparison}
\label{sssec:results-3}
We compared our method to an NMPC baseline which optimizes 
Problem~\ref{prob:nlp-specific} %
but in a receding horizon fashion similar to~\cite{fast_nonlin}. This baseline uses the same interior-point optimizer as our NLP stage for each of the receding horizon iterations, initializing each iteration with the previous result shifted by one step. 

To compare the merits of the two methods, we measured:
\begin{itemize}
    \item \textit{Solved}: percentage of dataset examples each method solved successfully (higher is better)
    \item \textit{Progress at $8s$} (P@8s): longitudinal distance covered along the reference path by the ego-vehicle within the first $8s$ of the example, computed over the set of solved examples (higher is better)
    \item \textit{Speed tracking} ($v$): average instantaneous speed in an example, computed over the set of solved examples (closer to the desired target of $8m/s$ is better)
    \item \textit{Absolute jerk value} ($|\dot{a}|$): average absolute value of longitudinal jerk in an example, computed over the set of solved examples (lower is better)
\end{itemize}

Absolute value of lateral jerk was not considered as part of our comparison since it can be misleading within the residential driving setup due to the fact that any overtaking maneuver will naturally incur higher lateral jerk values compared with situations when the planner does not perform an overtake, and thus it does not always reflect comfort. Longitudinal jerk, on the other hand, is invariant to the maneuver the planner is implementing, and can be considered a comfort metric.

\begin{table}[t!]
\centering
\bgroup
\def\arraystretch{1.4}
\begin{tabular}{l|c|c}
& \textsc{NMPC} & \textsc{Ours} \\ \Xhline{2\arrayrulewidth}
Solved (\%) & $87.79$ & $\mathbf{98.32}$ \\
Runtime (s) & $\mathbf{0.56}$ & $0.92$ \\ \hline
Progress (P@8s in $m$) & $46.24 \stddev{11.52}$ & $\mathbf{53.18} \stddev{12.87}$ \\
Speed ($v$ in $m/s$) & $6.00 \stddev{1.49}$ & $\mathbf{6.88} \stddev{1.65}$ \\
Jerk ($|\dot{a}|$ in $m/s^3$) & $0.47 \stddev{0.03}$ & $\mathbf{0.43} \stddev{0.05}$ \\ \Xhline{2\arrayrulewidth}
\end{tabular}
\egroup
\caption{\textit{NMPC Comparison}: percentage of dataset examples solved, planning runtime, and metrics measured over the set of examples solved by both methods (mean $\pm$ standard deviation) of progress after 8s (higher is better), average speed in an example (closer to $8m/s$ target is better), and average absolute value of jerk in an example (lower is better).}
\label{tab:nmpc-comp}
\vspace{-0.8em}
\end{table}

We generated a new dataset of 4000 examples, 1000 per driving scenario class, and solved them using our method and the NMPC baseline. 

Table~\ref{tab:nmpc-comp} shows the percentage of examples solved by the methods, as well as the aforementioned metrics for the subset of examples solved by both. Our method outperforms the baseline in terms of the number of solved examples by over $10\%$ (over 400 examples in total). On average, our method achieves significantly higher progress, better speed tracking across scenarios, and smaller longitudinal jerk values. This can also be seen from the distributions of the metric values over the set of examples solved by both methods presented in Fig.~\ref{fig:nmpc_comparison_distributions}. 
With respect to runtime, our implementation of the NMPC baseline using IPOPT took on average $0.56s$ to generate a solution to each problem, making it nearly 64\% faster than our method.

\section{Related Work}
\label{sec:related-work}

In this section we analyze works related to our method in the literature in terms of definitions of safety, vehicle dynamics modeling, and other applications of constrained optimization.

\subsection{Safety in Planning for Autonomous Driving}

Within the autonomous vehicles motion planning literature, an accepted definition of safety has been through the concept of the \emph{inevitable collision set} (ICS): the set of states which will inevitably lead to a collision (sometimes also referred to as regions of inevitable collision or target set), where guaranteeing safety can be translated into computing and avoiding the ICS~\cite{chan2008improved,mitchell2005time,bautin2010inevitable,althoff2010probabilistic,fraichard2004inevitable,blake2019fpr}. However, the computation of the ICS is intractable without some simplifying assumptions~\cite{chen2017correct}, which might lead to (i) undesirably-reactive behavior when the system approaches the boundary of the set and (ii) overly-conservative plans in interactive cases~\cite{fast_nonlin}. Similarly, reachability analysis methods, which compute the forward reachable set instead of the ICS for the sake of tractability, typically suffer from the same issues~\cite{althoff2014online, liu2017provably, wang2020infusing}. 

We follow the approach from~\cite{fast_nonlin}, in which safe regions are defined by probabilistic constraints on the ego states in the plan. While the chance constraints introduce a conservative factor in the planner when compared to assuming deterministic future positions, the departure from the strict ICS analysis creates an important gap between the executed future states of other agents and the uncertainty-aware predictions. In~\cite{fast_nonlin} the authors address this through a high re-planning rate: the frequent update of the planned trajectory as the true states of other agents are observed induces a low uncertainty at the level of the implemented actions at a certain time.

It should be noted that the notions of safety mentioned above are based on the fundamental assumption that the vehicle dynamics model used is representative of the full set of differential constraints of the system, and thus the generated plans are both safe and feasible in the real world~\cite{fast_nonlin,paden2016survey}. This is not always the case for low-fidelity formulations that use approximate dynamics and operate under model mismatch in which a high frequency feedback controller tracks the generated plan~\cite{singh2018robust}. Several works have shown that high-fidelity closed-form dynamics models can be used reliably in realistic simulation setups~\cite{andersen2017trajectory,fast_nonlin} and in the real world~\cite{verschueren2014towards,brown2017safe}. For all the experiments in this paper we used a kinematic bicycle model within the NLP problem (including in the NMPC implementation for Sec.~\ref{sssec:results-3}), yet the framework can be extended to higher fidelity models such as the one considered in~\cite{fast_nonlin}, possibly by considering a different linear vehicle model at the MILP stage (e.g. a first-order hold discretization as in~\cite{brown2017safe}).

\subsection{Motion Planning via Constrained Optimization}

Our framework fits within the constrained optimization literature~\cite{paden2016survey} using a receding horizon~\cite{laurense2019integrated,wurts2018collision}. While some related work tackles restricted settings of the problem using linear, mixed-integer linear, or convex program formulations~\cite{lima2015clothoid,babu2018model,schouwenaars2001mixed}, we directly solve a warm-started, nonlinear problem, handling nearly arbitrary scenarios with complex combinations of dynamic and static vehicles over a long horizon ($8s$ in the reported experiments). Rick~\textit{et ~al.}~\cite{rick2019autonomous} formulate an NLP problem similar to ours, but with no uncertainty over the predicted positions of road users at the optimization level, and solve it within an NMPC loop initialized with the shifted previous solution. While our planning formulation accounts for state uncertainty, the task of state estimation/filtering~\cite{albrecht2016exploiting} is outside the scope of this work.

The approach we take is closest to~\cite{fast_nonlin} in terms of the formulation of constraints (following from~\cite{liniger2015optimization, lam2012model, faulwasser2009model}), despite the fact that the problem in~\cite{fast_nonlin} refers to a corrective system in a semi-autonomous setting as opposed to the fully autonomous driving setting we tackle. Our work can be seen as a receding horizon approach, yet it differs from~\cite{fast_nonlin} in that the latter uses the shifted previous solution as the initialization for the NLP problem at each stage within an NMPC framework, whereas ours warm-starts each stage with a fresh MILP solution. Other works have used mixed-integer linear formulations for path planning~\cite{schouwenaars2001mixed}, but, to the best of our knowledge, this is the first work within the autonomous driving literature that uses mixed-integer linear formulations to warm-start an NLP solver.

\section{Discussion}
\label{sec:discussion}

Safety is paramount for autonomous driving. At deployment time, optimization-based planners require fallback mechanisms to deal with infeasible problems so that a plan is generated at each planning step. This is exacerbated for nonlinear non-convex problem settings as convergence is uncertain, even for feasible problems~\cite{fast_nonlin, local_optima_1}.
As such, a proxy for the deployed system safety could be seen in the percentage of solved problems. 
Our two-stage planner solves over 10\% more examples than the NMPC baseline in our scenario classes, implying that the fallback mechanism would have to be called significantly less frequently than in the case of that baseline. 

Furthermore, the presented results validate the claim that our framework with its informed initialization produces solutions of a higher quality than the ones produced by methods that rely on solving the NLP problem directly with simpler initializations or through an NMPC formulation that initializes it with a previous shifted solution. %

The analysis of the initialization ablation presented in Table~\ref{tab:init-comp-all} and Fig.~\ref{fig:init_boxplots} might suggest only marginal improvements by our method in some cases when looked at with the runtime increase (e.g., our solution outperforms \textsc{Zeros} by a margin of $2\%$ in terms of convergence and $11.62\%$ in terms of cost, but it runs on average $31\%$ slower). However, taking the sequential nature of the decision making problem into account -- i.e., the fact that the planning problems need to be solved continually instead of the one-shot planning reported in Table~\ref{tab:init-comp-all} -- means that these small differences will accumulate over longer runs.
This phenomenon is highlighted, for example,   by the receding horizon compounding in the NMPC comparison, as observed by the significant overall improvements in convergence and metrics in Table~\ref{tab:nmpc-comp} and Fig.~\ref{fig:nmpc_comparison_distributions}.

Generally, the non-convex nature of the NLP problem induces local optima that are highly influenced by the initialization, which, in the case of NMPC, is biased towards following a similar plan to the ones previously computed, and in the case of \textsc{Zeros}, is biased towards plans closer to the origin in the solution space. The MILP initialization we use, on the other hand, enables the subsequent NLP to explore different parts of the solution space when needed.

Finally, the initialization ablation over the formulation of Problem~\ref{prob:milp} strongly justifies the inclusion of collision avoidance and speed related constraints in our method, the effects of which can be observed by the increased convergence. As expected, removing these constraints reduces the complexity of the problem, resulting in better initialization runtime. However, the deterioration in the quality of the initialization is also clear from the increased NLP runtime, yielding a comparable total runtime to our method.

\section{Conclusion}
\label{sec:conclusion}

In this paper we introduce a two-stage optimization framework for autonomous driving which first solves a linearized version of the planning problem (formulated as a Mixed-Integer Linear Program in a receding horizon fashion) to warm-start the second nonlinear optimization stage. We show that our MILP initialization leads on average to a higher percentage of solved examples, lower costs, and better solving time for the NLP stage when compared with alternative initializations. Additionally, we show that the two-stage formulation solves more examples than an NMPC baseline, outperforming it in terms of progress and passenger comfort metrics.

By using generic, off-the-shelf solvers, our method trades off runtime for solution quality. One possible direction of future work is to investigate methods that can accelerate the solving time of the first stage, e.g.~\cite{milp_online,gupta2020hybrid,nair2020solving}, in order to to achieve faster runtime for real-world planning. This would also have the advantage of allowing for faster re-planning; bridging the safety gap between the uncertainty-aware predictions we use and the actual behavior of the other road users. Nonetheless, we have shown through our experiments that the current framework is still suitable for deployment in complex environments where the quality of the solution is extremely important.

\appendices

\begin{figure*}[t]
    \centering

    \begin{subfigure}[b]{0.255\textwidth}
        \centering
        \includegraphics[height=1.8cm]{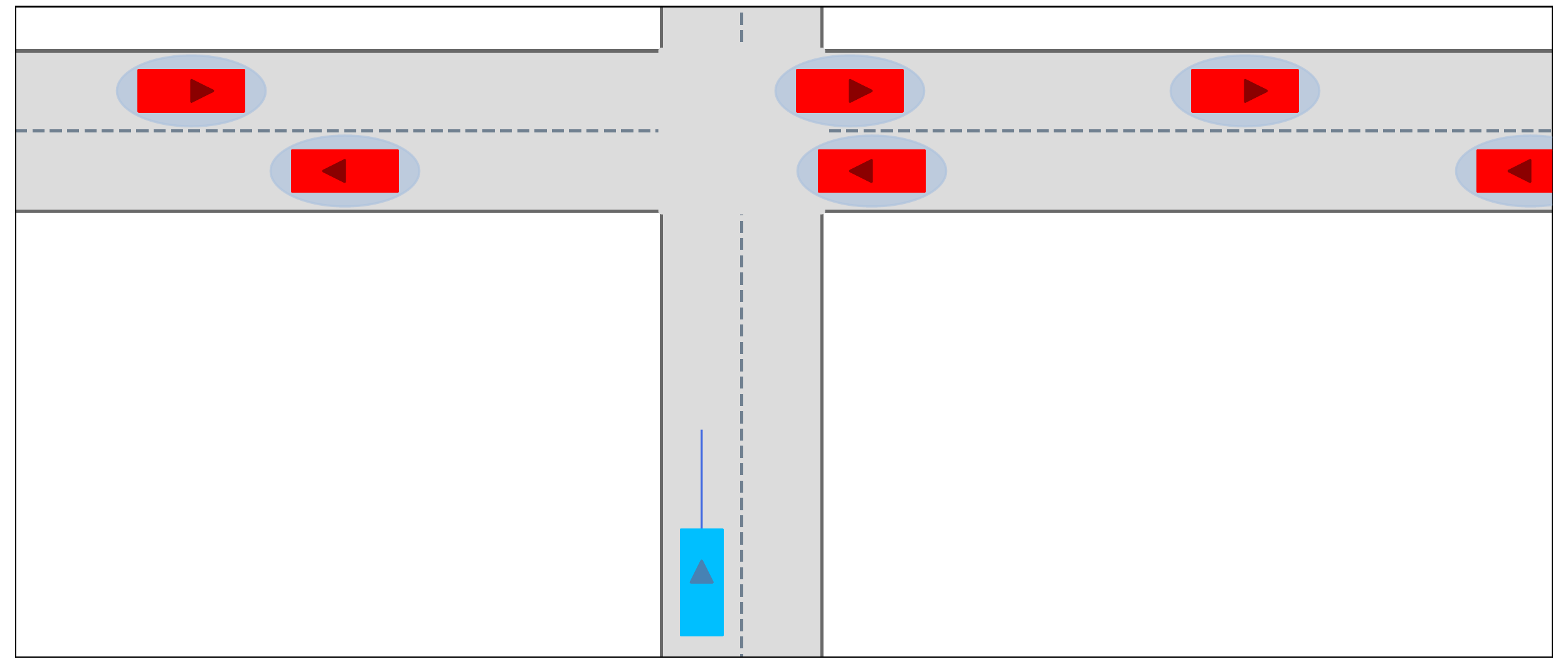}
        \caption{$t=0.0s$}
        \label{fig:busy_junction_0_0}
    \end{subfigure}
    \begin{subfigure}[b]{0.23\textwidth}
        \centering
        \includegraphics[height=1.8cm]{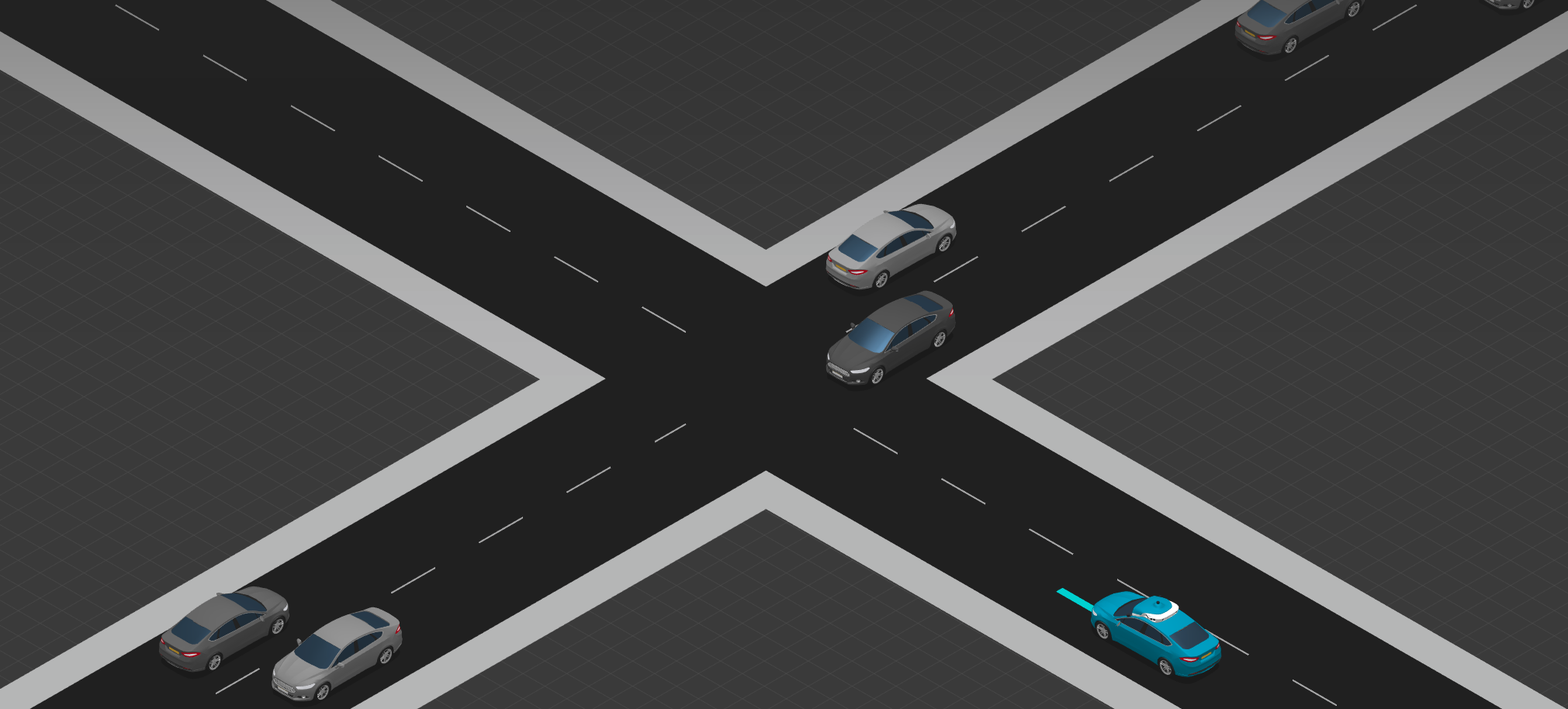}
        \caption{Simulator view ($t=0.47s$)}
        \label{fig:junction_sim_view}
    \end{subfigure}
    \begin{subfigure}[b]{0.24\textwidth}
        \centering
        \includegraphics[height=1.8cm]{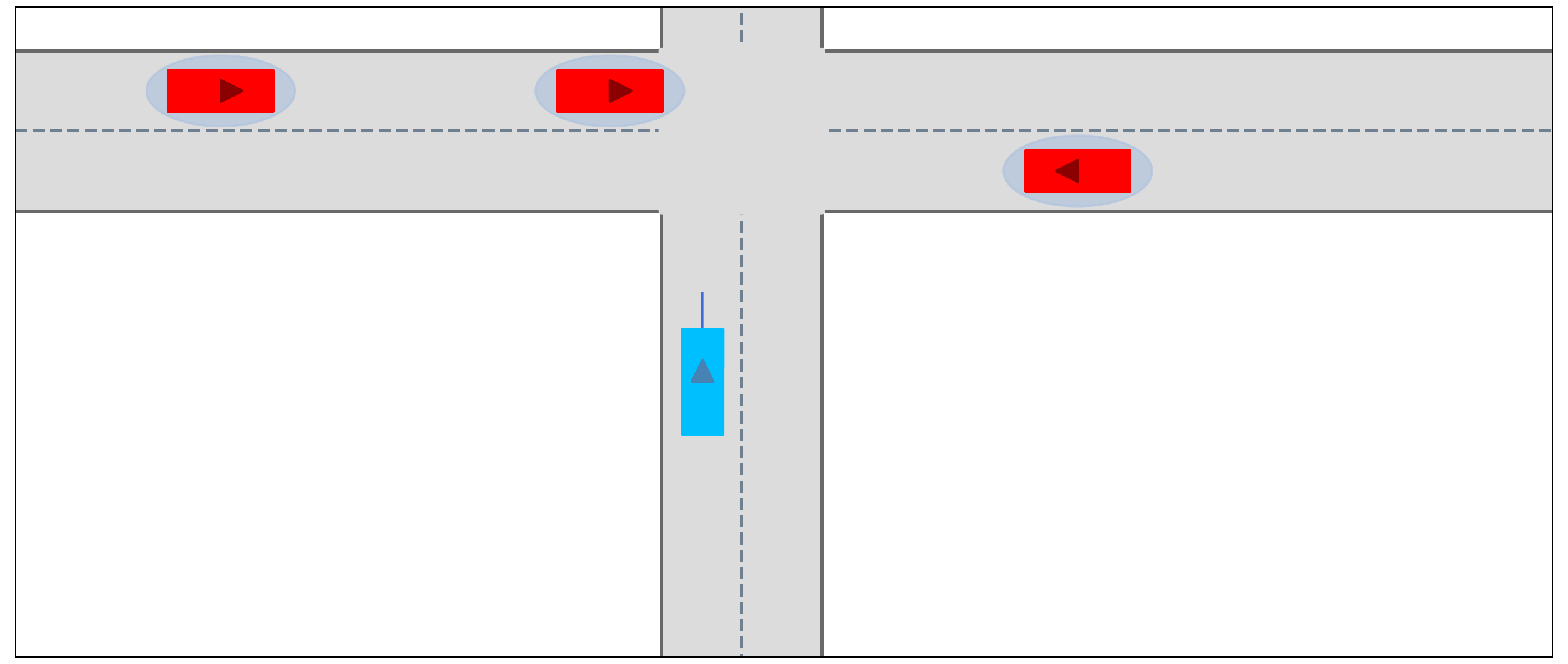}
        \caption{$t=6.47s$}
        \label{fig:busy_junction_3_2}
    \end{subfigure}
    \begin{subfigure}[b]{0.24\textwidth}
        \centering
        \includegraphics[height=1.8cm]{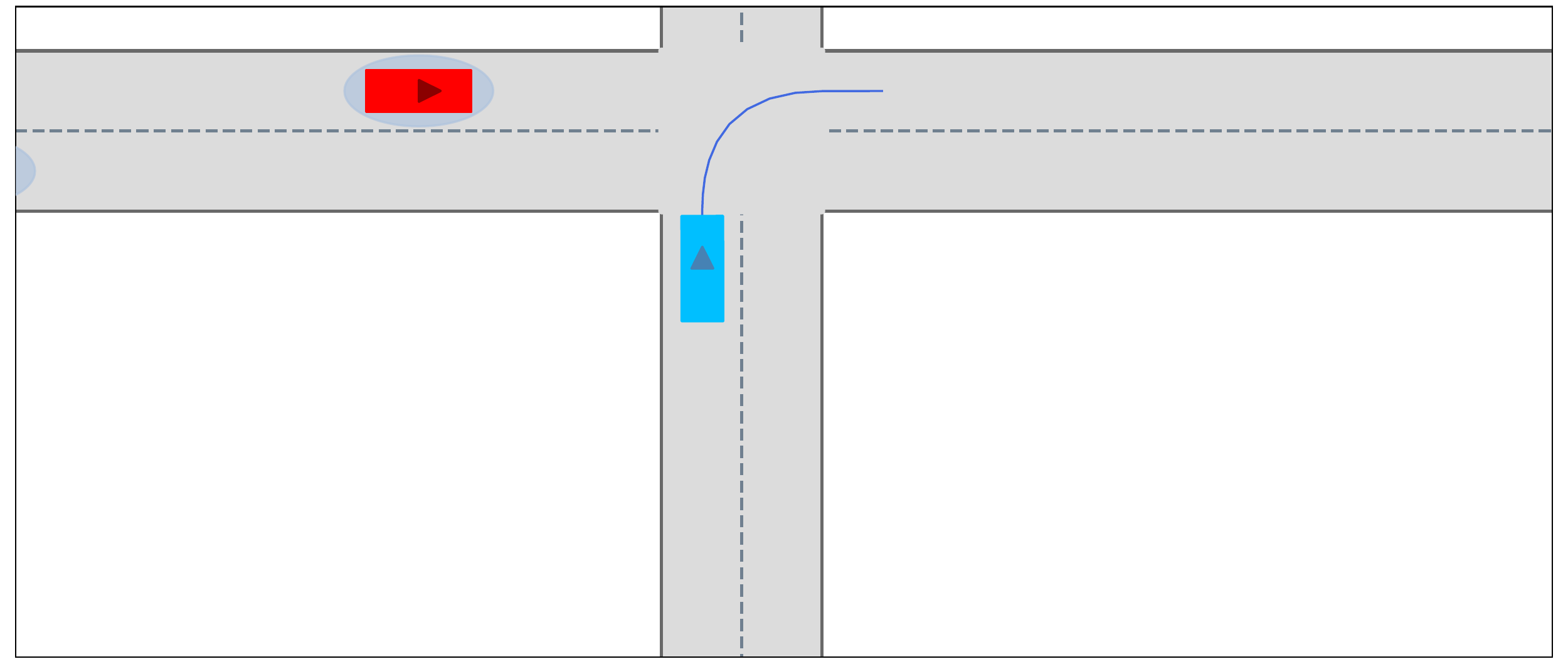}
        \caption{$t=13.47s$}
        \label{fig:busy_junction_6_07}
    \end{subfigure}
    \caption{Junction unprotected right turn example: (a), (c) and (d) showing the planner's view of a junction handling problem, with the ego-vehicle (blue), dynamic vehicles (red) and ego's plan (dark blue trace) at different times $t$; (b) shows our simulator's rendering of the situation at $t=0.47s$.
    }
    \label{fig:junction_three_plots}
    \vspace{-1.0em}
\end{figure*}

\section{Optimization parameters}
\label{sec:appendix-params}
\newcommand{\tablenameX}{TABLE A\hspace{-4pt}}
\renewcommand{\thetable}{\arabic{table}}
\setcounter{table}{0}

The parameters of Problem~\ref{prob:nlp-specific} and Problem~~\ref{prob:milp} in the context of the results in Sec.~\ref{sec:results} are defined in Table~\ref{tab:params}.

\begin{table}[H]
\centering
\bgroup
\def\arraystretch{1.3}
\begin{tabular}{lll||lll}
Parameter & Stage               & Value & Parameter & Stage               & Value \\ \Xhline{2\arrayrulewidth}
$L$& NLP & $4.8$ & $a^x_{\min}$ & MILP & $-3$ \\
$\delta_{\max}$ & NLP & $0.45$ & $a^x_{\max}$ & MILP & $3$\\
$a_{\min}$ & NLP & $-3$ & $a^y_{\min}$ & MILP & $-0.5$ \\
$a_{\max}$ & NLP & $3$ & $a^y_{\max}$ & MILP & $0.5$ \\
$\dot{a}_{\max}$ & NLP & $0.5$ & $\Delta a^x_{\max}$ & MILP & $0.5$ \\
$\dot{\delta}_{\max}$ & NLP & $0.18$ & $\Delta a^y_{\max}$ & MILP & $0.1$ \\
$\dot{a}_{\max}$ & NLP & $0.5$ & $v^x_{\min}$ & MILP & $0$ \\
$v_{\min}$ & NLP & $0$ & $v^x_{\max}$ & MILP & $3$ \\
$v_{\max}$ & NLP & $10$ & $v^y_{\min}$ & MILP & $-1$ \\
$\omega_x$ & NLP & $0.1$ & $v^y_{\max}$ & MILP & $1$ \\
$\omega_v$ & NLP & $2.5$ & $\Omega_x$ & MILP & $0.9$  \\
$\omega_y$ & NLP & $0.05$ & $\Omega_v$ & MILP & $0.5$ \\
$\omega_a$ & NLP & $1.0$ & $\Omega_y$ & MILP & $0.05$ \\
$\omega_{\delta}$ & NLP & $2.0$ & $\Omega_a$ & MILP & $0.4$ \\
$\rho$ & MILP & $1.5$ & $M$ & MILP & $10^4$ \\
$d$ & MILP & $0.9$ & & & \\ \Xhline{2\arrayrulewidth}
\end{tabular}
\egroup
\caption{Parameters used in the MILP and NLP optimization stages}
\label{tab:params}
\end{table}

\section{Junction Scenario}
\label{sec:appendix-junction}

Fig.~\ref{fig:junction_three_plots} shows the ego-vehicle navigating a junction and performing an unprotected right turn.

\section{Generation of dataset examples}
\label{sec:appendix-gen}
\renewcommand{\tablenameX}{TABLE B\hspace{-4pt}}
\renewcommand{\thetable}{\arabic{table}}
\setcounter{table}{0}

Here we give details of the procedural generation of scenarios presented in Sec.~\ref{ssec:scenarios-in-eval}. 

\begin{table}[H]
\centering
\def\arraystretch{1.3}
\begin{tabular}{ll|l|l}
\multicolumn{2}{l|}{Parameter}& Min & Max \\ \Xhline{2\arrayrulewidth}
\multicolumn{2}{l|}{Number of lanes} & $2$ & $2$ \\ 
\multicolumn{2}{l|}{Lane width ($m$)} & $3.5$ & $4.3$ \\\hline
Ego& initial $x$ ($m$) & $0$ & $0$ \\
& initial $y$ ($m$) & $b_l(x) + 0.55* 1.9$ & $b_r(x) - 0.55* 1.9$ \\
& initial $v$ ($ms^{-1}$) & $0$ & $9.5$ \\ 
& initial $\phi$ ($rad$) & $-\pi/12$ & $+\pi/12$ \\ \Xhline{2\arrayrulewidth}
\end{tabular}
\caption{Common parameters}
\label{tab:params-common}
\end{table}

We assume the ego-vehicle has length $4.8m$ and width $1.9m$, and that the scenario parameters are uniformly sampled from the ranges defined in Tables~\ref{tab:params-common}-\ref{tab:params-doov}.

\begin{table}[H]
\centering
\def\arraystretch{1.3}
\begin{tabular}{ll|l|l}
\multicolumn{2}{l|}{Parameter}& Min & Max \\ \Xhline{2\arrayrulewidth}
\multicolumn{2}{l|}{Number of static vehicles} & $2$ & $6$ \\ \hline
Static vehicle & $x$ ($m$) & $0$ & $80$ \\
& $y$ ($m$) & $b_l(x)$ & $b_r(x)$ \\
& width ($m$) & $1.7$ & $2.5$ \\
& length ($m$) & $4.0$ & $8.0$  \\ \Xhline{2\arrayrulewidth}
\end{tabular}
\caption{Parameters of scenario \texttt{SO}}
\label{tab:params-so}
\end{table}

\begin{table}[H]
\centering
\def\arraystretch{1.3}
\begin{tabular}{ll|l|l}
\multicolumn{2}{l|}{Parameter}& Min & Max \\ \Xhline{2\arrayrulewidth}
\multicolumn{2}{l|}{Number of static vehicles} & $2$ & $6$ \\ \hline
Static vehicle &  $x$ ($m$) & $0$ & $80$ \\
&  $y$ ($m$) & $b_l(x)$ & 0 \\
&  width ($m$) & $1.7$ & $2.5$  \\
&  length ($m$) & $4.0$ & $8.0$  \\ \hline
Oncoming vehicle & initial $x$ ($m$) & $20$ & $80$ \\
&  initial $y$ ($m$) & $b_r(x)/2$ & $b_r(x)/2$ \\
&  initial $v$ ($ms^{-1}$) & $1.0$ & $8.5$ \\
&  width ($m$) & $1.7$ & $2.5$ \\
&  length ($m$) & $4.0$ & $8.0$ \\ \Xhline{2\arrayrulewidth}
\end{tabular}
\caption{Parameters of scenario \texttt{SO+OV}}
\label{tab:params-soov}
\end{table}

\begin{table}[H]
\centering
\def\arraystretch{1.3}
\begin{tabular}{ll|l|l}
\multicolumn{2}{l|}{Parameter}& Min & Max \\ \Xhline{2\arrayrulewidth}
Dynamic vehicle & initial $x$ ($m$) & $20$ & $80$ \\
& initial $y$ ($m$) & $b_l(x)/2$ & $b_l(x)/2$ \\
& initial $v$ ($ms^{-1}$) & $0.5$ & $3.5$ \\
& width ($m$) & $1.7$ & $2.5$ \\
& length ($m$) & $4.0$ & $8.0$ \\ \Xhline{2\arrayrulewidth}
\end{tabular}
\caption{Parameters of scenario \texttt{DO}}
\label{tab:params-do}
\end{table}

\begin{table}[H]
\centering
\def\arraystretch{1.3}
\begin{tabular}{ll|l|l}
\multicolumn{2}{l|}{Parameter}& Min & Max \\ \Xhline{2\arrayrulewidth}
Oncoming vehicle & initial $x$ ($m$) & $20$ & $80$ \\
& initial $y$ ($m$) & $b_r(x)/2$ & $b_r(x)/2$ \\
& initial $v$ ($ms^{-1}$) & $1.0$ & $8.5$ \\
& width ($m$) & $1.7$ & $2.5$ \\
& length ($m$) & $4.0$ & $8.0$ \\ \hline
Dynamic vehicle & initial $x$ ($m$) & $20$ & $80$ \\
& initial $y$ ($m$) & $b_l(x)/2$ & $b_l(x)/2$ \\
& initial $v$ ($ms^{-1}$) & $0.5$ & $3.5$ \\
& width ($m$) & $1.7$ & $2.5$ \\
& length ($m$) & $4.0$ & $8.0$ \\ \Xhline{2\arrayrulewidth}
\end{tabular}
\caption{Parameters of scenario \texttt{DO+OV}}
\label{tab:params-doov}
\end{table}

\bibliographystyle{IEEEtran}
\bibliography{refs}

\begin{IEEEbiography}[{\includegraphics[width=1in,height=1.25in,clip,keepaspectratio]{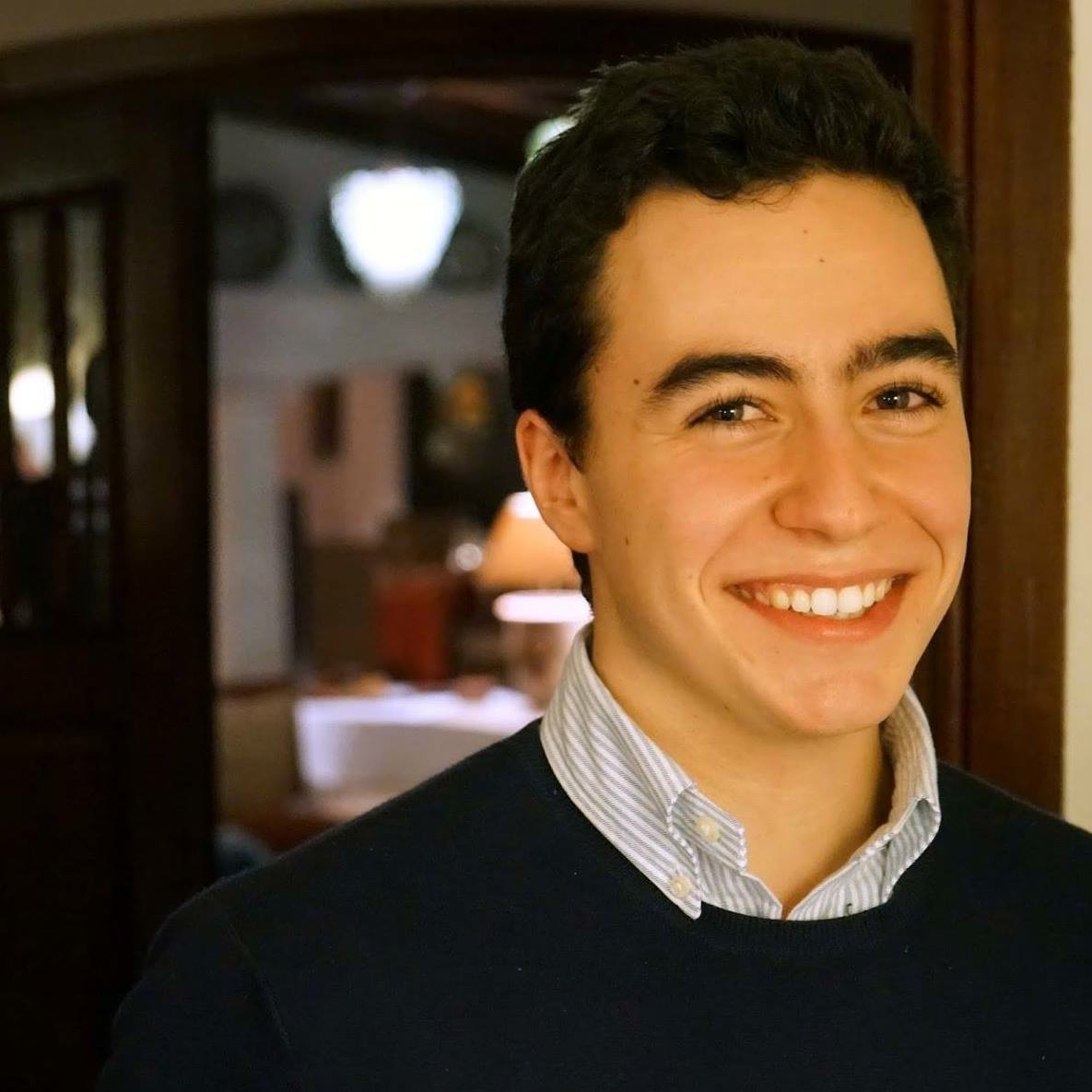}}]{Francisco Eiras}
received a BSc in Electrical and Computer Engineering from Tecnico Lisbon in 2016, and an MSc in Computer Science from the University of Oxford in 2018. He worked as a Research Engineer within the Motion Planning and Prediction Applied Research team at Five for 2 years following his MSc, and is currently pursuing a PhD in Engineering Science at the Optimization for Vision and Learning (OVAL) group at the University of Oxford.
\end{IEEEbiography}

\begin{IEEEbiography}[{\includegraphics[width=1in,height=1.25in,clip,keepaspectratio]{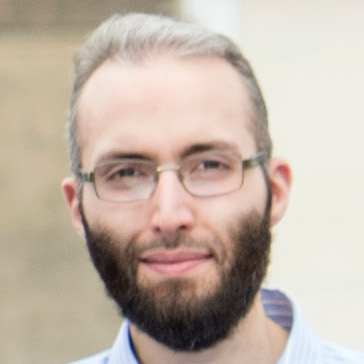}}]{Majd Hawasly} is a Senior Research Scientist in the Motion Planning and Prediction Applied Research team at Five. He received his PhD from the School of Informatics at the University of Edinburgh in 2014. Prior to his role at Five, he was a postdoctoral research fellow at the University of Leeds.
\end{IEEEbiography}

\begin{IEEEbiography}[{\includegraphics[width=1in,height=1.25in,clip,keepaspectratio]{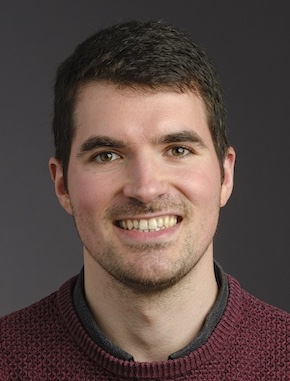}}]{Stefano V. Albrecht} is Assistant Professor in Artificial Intelligence in the School of Informatics, University of Edinburgh, where he leads the Autonomous Agents Research Group (https://agents.inf.ed.ac.uk). He is a Royal Society Industry Fellow working with UK-based company Five to develop AI technologies for autonomous vehicles. His research interests are in the areas of autonomous agents, multi-agent interaction, reinforcement learning, and game theory, with a focus on sequential decision making under uncertainty. Previously, he was a postdoctoral fellow at the University of Texas at Austin. He obtained PhD/MSc degrees from the University of Edinburgh and a BSc degree from Technical University of Darmstadt.
\end{IEEEbiography}

\vspace{-12.5cm}

\begin{IEEEbiography}
    [{\includegraphics[width=1in,height=1.25in,clip,keepaspectratio]{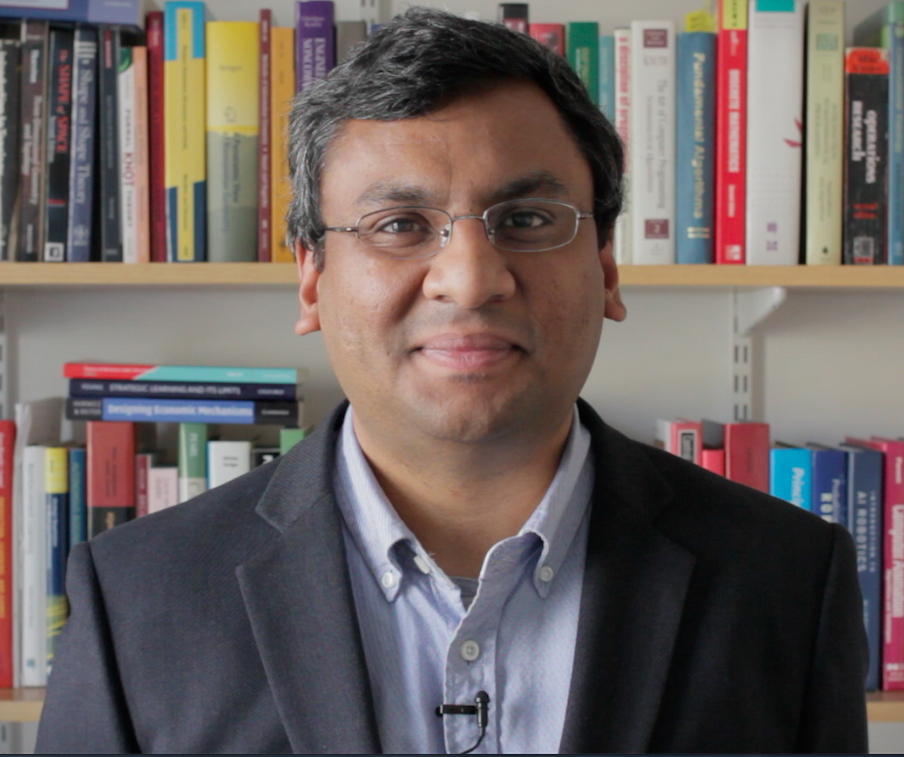}}]
    {Subramanian Ramamoorthy}
is a Professor in the School of Informatics at the University of Edinburgh, where he holds the Personal Chair of Robot Learning and Autonomy. He is a Turing Fellow at the Alan Turing Institute, Executive Committee Member for the Edinburgh Centre for Robotics, and a Member of the UK Computing Research Committee. He received his PhD in Electrical and Computer Engineering from The University of Texas at Austin in 2007. He has been a Member of the Young Academy of Scotland at the Royal Society of Edinburgh, and has held Visiting Professor positions at the University of Rome ``La Sapienza'' and at Stanford University. His research investigates learning, adaptation, and control mechanisms that enable autonomous robots to cope with significant uncertainty and changes in tasks and environments. 
Between 2017-2020, he served as Vice-President -- Prediction and Planning -- at Five. He continues to be involved with the company as a Scientific Advisor. %
\end{IEEEbiography}

\end{document}